\documentclass{article}

\usepackage[compress]{natbib}
\usepackage[resetlabels]{multibib}
\usepackage[linesnumbered,ruled,vlined]{algorithm2e}
\newcites{APP}{References}

\usepackage[utf8]{inputenc} 
\usepackage[T1]{fontenc}    
\usepackage{float}
\usepackage[table]{xcolor}
\definecolor{dred}{rgb}{0.6,0,0}
\definecolor{dpurple}{HTML}{A020F0}
\definecolor{dblue}{rgb}{0,0,0.6}
\definecolor{hlcolor}{rgb}{1,1,0.8}
 
\usepackage[colorlinks=true, linkcolor=dblue, citecolor=dred]{hyperref}       

\usepackage{graphicx}
\usepackage{caption}
\usepackage{subcaption}
\usepackage{amsmath,amssymb,bm}

\usepackage[nameinlink]{cleveref}
\usepackage{url}            
\usepackage{booktabs}       
\usepackage{amsfonts}       
\usepackage{nicefrac}       
\usepackage{microtype}      
\usepackage{soul}
\usepackage{authblk}
\usepackage{lmodern}
\usepackage{tabularx}
\usepackage{braket}
\usepackage[margin=1in]{geometry}
\newcolumntype{Y}{>{\centering\arraybackslash}X}

\usepackage[page]{appendix}

\sethlcolor{hlcolor}
\Crefname{equation}{Equation}{Equations}
\Crefname{figure}{Figure}{Figures}
\creflabelformat{equation}{#2#1#3}
\crefrangelabelformat{equation}{#3#1#4-#5#2#6}

\definecolor{blue}{HTML}{0000FF}
\definecolor{green}{HTML}{008000}
\definecolor{cyan}{HTML}{00BFBF}
 
\renewcommand\b\bm

\newcommand{\vect}{{\text{vec}}}
\DeclareMathOperator*{\argmax}{arg\,max}
\DeclareMathOperator*{\argmin}{arg\,min}
\newcommand{\CrefNoLink}[1]{\Cref{#1}}

\title{Natural continual learning:\\success is a journey, not (just) a destination}

\author[1]{\normalsize \bfseries Ta-Chu Kao${}^{*\rm @}$}
\author[1]{\bfseries Kristopher T.\ Jensen${^{*}}$}
\author[1,2]{\bfseries Gido M. van de Ven}
\author[3]{{}\\ \bfseries Alberto Bernacchia}
\author[1]{\bfseries Guillaume Hennequin}
\affil[1]{ \small Computational and Biological Learning Lab, Department of Engineering, University of Cambridge, Cambridge, UK}
\affil[2]{ \small Center for Neuroscience and Artificial Intelligence, Baylor College of Medicine}
\affil[3]{ \small MediaTek Research, Cambridge, UK}
\date{\vspace*{-1.5em}
\normalsize ${}^{*}$ These authors contributed equally \hspace{1em}
\normalsize ${}^{\rm @}$ Corresponding author (tck29@cam.ac.uk)}

\begin{document}

\maketitle

\begin{abstract}
\noindent
Biological agents are known to learn many different tasks over the course of their lives, and to be able to revisit previous tasks and behaviors with little to no loss in performance.
In contrast, artificial agents are prone to `catastrophic forgetting' whereby performance on previous tasks deteriorates rapidly as new ones are acquired.
This shortcoming has recently been addressed using methods that encourage parameters to stay close to those used for previous tasks.
This can be done by (i) using specific parameter regularizers that map out suitable destinations in parameter space, or (ii) guiding the optimization journey by projecting gradients into subspaces that do not interfere with previous tasks.
However, these methods often exhibit subpar performance in both feedforward and recurrent neural networks, with recurrent networks being of interest to the study of neural dynamics supporting biological continual learning. 
In this work, we propose Natural Continual Learning (NCL), a new method that unifies weight regularization and projected gradient descent.
NCL uses Bayesian weight regularization to encourage good performance on all tasks at convergence and combines this with gradient projection using the prior precision, which prevents catastrophic forgetting during optimization.
Our method outperforms both standard weight regularization techniques and projection based approaches when applied to continual learning problems in feedforward and recurrent networks.
Finally, the trained networks evolve task-specific dynamics that are strongly preserved as new tasks are learned, similar to experimental findings in biological circuits.
\end{abstract}

\section{Introduction}
\label{sec:intro}

Catastrophic forgetting is a common feature of machine learning algorithms, where training on a new task often leads to poor performance on previously learned tasks.
This is in contrast to biological agents which are capable of learning many different behaviors over the course of their lives with little to no interference across tasks.
The study of continual learning in biological networks may therefore help inspire novel approaches in machine learning, while the development and study of continual learning algorithms in artificial agents can help us better understand how this challenge is overcome in the biological domain.
This is particularly true for more challenging continual learning settings where task identity is not provided at test time, and for continual learning in recurrent neural networks (RNNs), which is important due to the practical and biological relevance of RNNs.
However, continual learning in these settings has recently proven challenging for many existing algorithms, particularly those that rely on parameter regularization to mitigate forgetting \citep{ehret2020continual,duncker2020organizing,van2019three}.
In this work, we address these shortcomings by developing a continual learning algorithm that not only encourages good performance across tasks at convergence but also regularizes the optimization path itself using trust region optimization. 
This leads to improved performance compared to existing methods.

Previous work has addressed the challenge of continual learning in artificial agents using weight regularization, where parameters important for previous tasks are regularized to stay close to their previous values \citep{aljundi2018memory,kirkpatrick2017overcoming,huszar2017quadratic,nguyen2017variational,ritter2018online,zenke2017continual}.
This approach can be motivated by findings in the neuroscience literature of increased stability for a subset of synapses after learning \citep{xu2009rapid, yang2009stably}.
More recently, approaches based on projecting gradients into subspaces orthogonal to those that are important for previous tasks have been developed in both feedforward \citep{zeng2019continual, saha2021gradient} and recurrent \citep{duncker2020organizing} neural networks.
This is consistent with experimental findings that neural dynamics often occupy orthogonal subspaces across contexts in biological circuits \citep{kaufman2014cortical, ames2019motor,failor2021learning,jensen2021scalable}.
While these methods have been found to perform well in many continual learning settings, they also suffer from several shortcomings.
In particular, while Bayesian weight regularization provides a natural way to weigh previous and current task information, this approach can fail in practice due to its approximate nature and often requires additional tuning of the importance of the prior beyond what would be expected in a rigorous Bayesian treatment \citep{van2018generative}.
In contrast, while projection-based methods have been found empirically to mitigate catastrophic forgetting, it is unclear how the `important subspaces' should be selected and how such methods behave when task demands begin to saturate the network capacity.

In this work, we develop natural continual learning (NCL), a new method that combines (i) Bayesian continual learning using weight regularization with (ii) an optimization procedure that relies on a trust region constructed from an approximate posterior distribution over the parameters given previous tasks.
This encourages parameter updates predominantly in the null-space of previously acquired tasks while maintaining convergence to a maximum of the Bayesian approximate posterior.
We show that NCL outperforms previous continual learning algorithms in both feedforward and recurrent networks.
We also show that the projection-based methods introduced by \citet{duncker2020organizing} and \citet{zeng2019continual} can be viewed as approximations to such trust region optimization using the posterior from previous tasks.
Finally, we use tools from the neuroscience literature to investigate how the learned networks overcome the challenge of continual learning.
Here, we find that the networks learn latent task representations that are stable over time after initial task learning, consistent with results from biological circuits.

\section{Method}
\label{sec:method}

\paragraph{Notations} We use  $\b{X}^\top$, $\b{X}^{-1}$, $\text{Tr}(\b{X})$ and $\text{vec}(\b{X})$ to denote the transpose, inverse, trace, and column-wise vectorization of a matrix $\b{X}$.
We use $\b{X} \otimes \b{Y}$ to represent the Kronecker product between matrices $\b{X} \in \mathbb{R}^{n \times n}$ and $\b{Y} \in \mathbb{R}^{m \times m}$ such that $(\b{X} \otimes \b{Y})_{mi + k ,m j + l }  = \b{X}_{ij} \b{Y}_{kl}$.
We use bold lower-case letters $\b{x}$ to denote column vectors.
$\mathcal{D}_k$ refers to a `dataset' corresponding to task $k$, which in this work generally consists of a set of input-output pairs $\{ \b{x}_k^{(i)}, \b{y}_k^{(i)} \}$ such that $\ell_k(\b{\theta}) := \log p(\mathcal{D}_k | \b{\theta}) = \sum_i \log p_{\b\theta}(\b{y}^{(i)}_k | \b{x}^{(i)}_k)$ is the task-related performance on task $k$ for a model with parameters $\b{\theta}$.
Finally, we use $\hat{\mathcal{D}}_k$ to refer to a dataset generated by inputs from the $k^{th}$ task where $ \{ \hat{\b{y}}_k^{(i)} \sim p_{\b\theta}(\b{y} | \b{x}^{(i)}_k) \}$ are drawn from the model distribution $\mathcal{M}$.

\subsection{Bayesian continual learning}
\label{subsec:bcl}

\paragraph{Problem statement}
In continual learning, we train a model on a set of $K$ tasks $\{ \mathcal{D}_1, \ldots, \mathcal{D}_K \}$ that arrive sequentially, where the data distribution $\mathcal{D}_k$ for task $k$ in general differs from $\mathcal{D}_{\neq k}$.
The aim is to learn a probabilistic model $p(\mathcal{D}|\b{\theta})$ that performs well on all tasks.
The challenge in the continual learning setting stems from the sequential nature of learning, and in particular from the common assumption that the learner does not have access to ``past'' tasks (i.e., $\mathcal{D}_j$ for $j < k$) when learning task $k$.
While we enforce this stringent condition in this paper, our approach may be easily combined with memory-based techniques such as coresets or generative replay \citep{ehret2020continual,von2019continual,nguyen2017variational,pan2020continual,shin2017continual,van2020brain,cong2020gan,rebuffi2017icarl,de2020continual,rolnick2019experience,titsias2020functional}.

\paragraph{Bayesian approach}
The continual learning problem is naturally formalized in a Bayesian framework whereby the posterior after $k-1$ tasks is used as a prior for task $k$.
More specifically, we choose a prior $p(\b{\theta})$ on the model parameters and compute the posterior after observing $k$ tasks according to Bayes' rule:
\begin{align}
	p(\b{\theta} | \mathcal{D}_{1:k}) \nonumber
	 & \propto p(\b{\theta}) \prod_{k'=1}^k p(\mathcal{D}_{k'}|\b{\theta})                         \\
	 & \propto p(\b{\theta}| \mathcal{D}_{1:k-1})  p(\mathcal{D}_k|\b{\theta}), \label{eq:cl_post}
\end{align}
where $\mathcal{D}_{1:k}$ is a concatenation of  the first $k$ tasks $(\mathcal{D}_1, \ldots, \mathcal{D}_k)$.
In theory, it is thus possible to compute the exact posterior $p(\b{\theta}|\mathcal{D}_{1:k})$ after $k$ tasks, while only observing $\mathcal{D}_k$, by using the posterior $p(\b{\theta}|\mathcal{D}_{1:k-1})$ after $k-1$ tasks as a prior.
However, as is often the case in Bayesian inference, the difficulty here is that the posterior is typically intractable.
To address this challenge, it is common to perform approximate online Bayesian inference.
That is, the posterior $p(\b{\theta}|\mathcal{D}_{1:k-1})$ is approximated by a parametric distribution with parameters $\b{\phi}_{k-1}$.
The approximate posterior $q(\b{\theta}; \b{\phi}_{k-1})$ is then used as a prior for task $k$.

\paragraph{Online Laplace approximation}
A common approach is to use the Laplace approximation whereby the posterior $p(\b{\theta}|\mathcal{D}_{1:k-1})$ is approximated as a multivariate Gaussian $q$ using local gradient information \citep{kirkpatrick2017overcoming,ritter2018online,huszar2017quadratic}.
This involves (i) finding a mode $\b{\mu}_{k}$ of the posterior during task $k$, and (ii) performing a second-order Taylor expansion around $\b{\mu}_k$ to construct an approximate Gaussian posterior
$q(\b{\theta}; \b{\phi}_{k}) = \mathcal{N}(\b{\theta}; \b{\mu}_{k}, \b{\Lambda}_k^{-1})$,
where $\b{\Lambda}_k$ is the precision matrix and $\b{\phi}_k = (\b{\mu}_k, \b{\Lambda}_k)$.
In this case, gradient-based optimization is used to find the posterior mode on task $k$ (c.f. \Cref{eq:cl_post}):
\begin{align}
	\b{\mu}_k
	 & = \argmax_{\b{\theta}}~
	\log p(\b{\theta}|\mathcal{D}_k, \b{\phi}_{k-1})         \\
	 & = \argmax_{\b{\theta}}~
	\log p(\mathcal{D}_k | \b{\theta}) +
	\log q(\b{\theta};\b{\phi}_{k-1})
	\\
	 & = \argmax_{\b{\theta}}~ \underbrace{\ell_k(\b\theta)-
		\frac{1}{2} (\b{\theta} - \b{\mu}_{k-1})^{\top}\b{\Lambda}_{k-1}(\b{\theta} - \b{\mu}_{k-1})}_{\displaystyle := \mathcal{L}_k(\b\theta)}
	\label{eq:laplace_update_mu}
\end{align}
The precision matrix $\b{\Lambda}_k$ is given by the Hessian of the negative log posterior at $\b{\mu}_k$:
\begin{align}
	\b{\Lambda}_k
	=
	-\left .
	\nabla_{\b{\theta}}^2 \log p(\b{\theta}|\mathcal{D}_k, \b{\phi}_{k-1})
	\right |_{\b{\theta} = \b{\mu}_k}
	=
	H(\mathcal{D}_k, \b{\mu}_k)
	+
	\b{\Lambda}_{k-1},
	\label{eq:laplace_update_lambda}
\end{align}
where $H(\mathcal{D}_k, \b{\mu}_k) =
	-
	\left .
	\nabla_{\b{\theta}}^2 \log p(\mathcal{D}_k | \b{\theta})
	\right |_{\b{\theta} = \b{\mu}_k}$ is the Hessian of the negative log likelihood of $\mathcal{D}_k$.

Continual learning with the online Laplace approximation thus involves two steps for each new task $\mathcal{D}_k$.
First, given $\mathcal{D}_k$ and the previous posterior $q(\b{\theta};\b{\mu}_{k-1}, \b{\Lambda}_{k-1}^{-1})$ (i.e. the new prior), $\b{\mu}_{k}$ is found using gradient-based optimization (\Cref{eq:laplace_update_mu}).
This step can be interpreted as optimizing the likelihood of $\mathcal{D}_k$ while penalizing changes in the parameters $\b{\theta}$ according to their importance for previous tasks, as determined by the prior precision matrix $\b{\Lambda}_{k-1}$.
Second, the new posterior precision matrix $\b{\Lambda}_k$ is computed according to \Cref{eq:laplace_update_lambda}.

\paragraph{Approximating the Hessian}
In practice, computing $\b{\Lambda}_k$ presents two major difficulties.
First, because $q(\b{\theta} ; \b{\phi}_k)$ is a Gaussian distribution, $\b{\Lambda}_k$ has to be positive semi-definite (PSD), which is not guaranteed for the Hessian $H(\mathcal{D}_k, \b{\mu}_k)$.
Second, if the number of model parameters $n_\theta$ is large, it may be prohibitive to compute a full $( n_\theta \times n_\theta)$ matrix.
To address the first issue, it is common to approximate the Hessian with the Fisher information matrix (FIM; \citealp{martens2014new,huszar2017quadratic,ritter2018online}):
\begin{align}
	\b{F}_k =
	\left .
	\mathbb{E}_{p(\hat{\mathcal{D}}_k|\b{\theta})}
	\left [
		\nabla_{\b{\theta}} \log p(\hat{\mathcal{D}}_k|\b{\theta})
		\nabla_{\b{\theta}} \log p(\hat{\mathcal{D}}_k|\b{\theta})^{\top}
		\right ]
	\right |_{\b{\theta}=\b{\mu}_k}
	\approx H(\mathcal{D}_k, \b{\mu}_k)
	\label{eq:fisher_k}
\end{align}
The FIM is PSD, which ensures that $\b{\Lambda}_k = \sum_{k'=1}^k \b{F}_{k'}$ is also PSD.
Computing $\b{F}_k$ may still be impractical if there are many model parameters, and it is therefore common to further approximate the FIM using structured approximations with fewer parameters.
In particular, a diagonal approximation to $\b{F}_k$ recovers Elastic Weight Consolidation (EWC; \citealp{kirkpatrick2017overcoming}), while a Kronecker-factored approximation~\citep{martens2015optimizing} recovers the method proposed by \citet{ritter2018online}.
We denote this method `KFAC' and use it in \Cref{sec:exps} as a comparison for our own Kronecker-factored method.

\subsection{Natural continual learning}
\label{subsec:ngcl}

\begin{figure}[!t]
    \centering
    \includegraphics[width = 0.99\textwidth, trim={0 0 0 0}, clip=true]{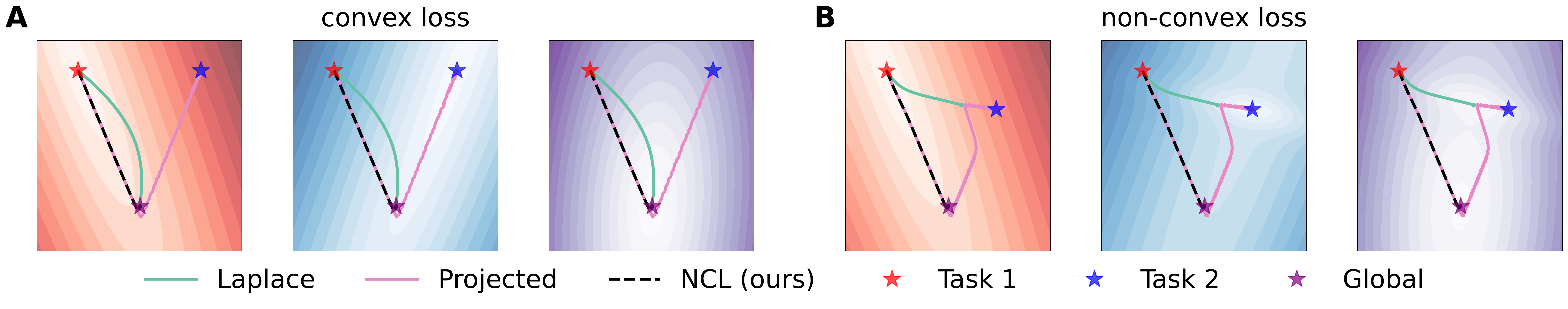}
    \caption{\label{fig:schematic}
        {\bfseries Continual learning in a toy problem.}
        {\bfseries (A)}~Loss landscapes of task 1 ($\ell_1$; left), task 2 ($\ell_2$; middle) and the combined loss $\ell_{1+2} = \ell_1 + \ell_2$ (right).
        Stars indicate the global optima for $\ell_1$ (red), $\ell_2$ (blue), and $\ell_{1+2}$ (purple).
        We assume that $\b{\theta}$ has been optimized for $\ell_1$ and consider how learning proceeds on task 2 using either the Laplace posterior (`Laplace', green), projected gradient descent on $\ell_2$ with preconditioning according to task 1 (`Projected', pink), or NCL (black dashed).
        Laplace follows the steepest gradient of $\ell_{1+2}$ and transiently forgets task 1.
        NCL follows a flat direction of $\ell_1$ and converges to the global optimum of $\ell_{1+2}$ with good performance on task 1 throughout.
        Projected gradient descent follows a similar optimization path to NCL but eventually diverges towards the optimum of $\ell_2$.
        {\bfseries (B)}~As in (A), now with non-convex $\ell_2$ (center), leading to a second local optimum of $\ell_{1+2}$ (right) while $\ell_1$ is unchanged (left).
        In this case, Laplace can converge to a local optimum which has `catastrophically' forgotten task 1.
        Projected gradient descent moves only slowly in `steep' directions of $\ell_1$ but eventually converges to a minimum of $\ell_2$.
        Finally, NCL finds a local optimum of $\ell_{1+2}$ which retains good performance on task 1.
        See \CrefNoLink{sec:toy} for further mathematical details.
        }
    \vspace*{-1em}
\end{figure}

While the online Laplace approximation has been applied successfully in several continual learning settings \citep{kirkpatrick2017overcoming,ritter2018online}, it has also been found to perform sub-optimally on a range of problems \citep{van2018generative,duncker2020organizing}.
Additionally, its Bayesian interpretation in theory prescribes a unique way of weighting the contributions of previous and current tasks to the loss.
However, to perform well in practice, weight regularization approaches have been found to require ad-hoc re-weighting of the prior term by several orders of magnitude \citep{kirkpatrick2017overcoming,ritter2018online,van2018generative}.
These shortcomings could be due to an inadequacy of the approximations used to construct the posterior (\Cref{subsec:bcl}).
However, we show in \Cref{fig:schematic} that standard gradient descent on the Laplace posterior has important drawbacks even in the exact case.
First, we show that exact Bayesian inference on a simple continual regression problem can produce indirect optimization paths along which previous tasks are transiently forgotten as a new task is being learned (\Cref{fig:schematic}A; green).
Second, when the loss is non-convex, we show that exact Bayesian inference can still lead to catastrophic forgetting (\Cref{fig:schematic}B; green).

An alternative approach that has found recent success in a continual learning setting involves projection based methods which restrict parameter updates to a subspace that does not interfere with previous tasks \citep{zeng2019continual,duncker2020organizing}.
However, it is not immediately obvious how this projected subspace should be selected in a way that appropriately balances learning on previous and current tasks.
Additionally, such projection-based algorithms have fixed points that are minima of the current task, but not necessarily minima of the (negative) Bayesian posterior.
This can lead to catastrophic forgetting in the limit of long training times (\Cref{fig:schematic}; pink), unless the learning rate is exactly zero in directions that interfere with previous tasks.

To combine the desirable features of both classes of methods, we introduce ``Natural Continual Learning'' (NCL) -- an extension of the online Laplace approximation that also restricts parameter updates to directions which do not interfere strongly with previous tasks.
In a Bayesian setting, we can conveniently express what is meant by such directions in terms of the prior precision matrix $\b{\Lambda}$.
In particular, `flat' directions of the prior (low precision) correspond to directions that will not significantly affect the performance on previous tasks.
Formally, we derive NCL as the solution of a trust region optimization problem.
This involves minimizing the posterior loss $\mathcal{L}_k(\b{\theta})$ within a region of radius $r$ centered around $\b{\theta}$ with a distance metric of the form $d(\b{\theta}, \b{\theta}+\b{\delta}) = \sqrt{\b{\delta}^\top \b{\Lambda}_{k-1} \b{\delta} /2}$ that takes into account the curvature of the prior via its precision matrix $\b{\Lambda}_{k-1}$:
\begin{align}
	 & \b{\delta}
	=
	\argmin_{\b{\delta}}
	\mathcal{L}_k(\b{\theta})
	+
	\nabla_{\b{\theta}} \mathcal{L}_k(\b{\theta})^\top
	\b{\delta}
	\quad
	\text{subject to}
	\;\;
	\frac{1}{2}\b{\delta}^\top \b{\Lambda}_{k-1} \b{\delta} \leq r^2,
	\label{eq:ncl_trust_region}
\end{align}
where
$
	\mathcal{L}_k(\b{\theta} + \b{\delta}) \approx
	\mathcal{L}_k(\b{\theta}) +
	\nabla_{\b{\theta}} \mathcal{L}_k(\b{\theta})^\top \b{\delta}$
is a first-order approximation to the updated Laplace objective.
The solution to this subproblem is given by
$\b{\delta} \propto \b{\Lambda}_{k-1}^{-1} \nabla_{\b{\theta}} \ell_k(\b{\theta}) - (\b{\theta} - \b{\mu}_{k-1})$ (see \CrefNoLink{sec:alt_ncl} for a derivation),
which gives rise to the NCL update rule
\begin{equation}
	\b{\theta} \leftarrow \b{\theta}
	+\gamma
	\left [
		\b{\Lambda}_{k-1}^{-1} \nabla_{\b{\theta}}
		\ell_k(\b{\theta})
		-
		(\b\theta - \b{\mu}_{k-1})
		\right ]
	\label{eq:ncl_update}
\end{equation}
for a learning rate parameter $\gamma$ (which is implicitly a function of $r$ in \Cref{eq:ncl_trust_region}).
To get some intuition for this learning rule, we note that $\b{\Lambda}_{k-1}^{-1}$ acts as a preconditioner for the first (likelihood) term, which drives learning on the current task while encouraging parameter changes predominantly in directions that do not interfere with previous tasks.
Meanwhile, the second term encourages $\b{\theta}$ to stay close to $\b{\mu}_{k-1}$, the optimal parameters for the previous task.
As we illustrate in \Cref{fig:schematic}, this combines the desirable features of both Bayesian weight regularization and projection-based methods.
In particular, NCL shares the fixed points of the Bayesian posterior while also mitigating intermediate or complete forgetting of previous tasks by preconditioning with the prior covariance.
Notably, if the loss landscape is non-convex (as it generally will be), NCL can converge to a different local optimum from standard weight regularization despite having the same fixed points (\Cref{fig:schematic}B).

\paragraph{Implementation}
The general NCL framework can be applied with different approximations to the Fisher matrix $\b{F}_k$ in \Cref{eq:fisher_k}~(see \Cref{subsec:bcl}).
In this work, we use a Kronecker-factored approximation \citep{martens2015optimizing,ritter2018online}.
However, even after making a Kronecker-factored approximation to $\b{F}_k$ for each task $k$, it remains difficult to compute the inverse of a sum of $k$ Kronecker products (c.f.\ \Cref{eq:laplace_update_lambda}).
To address this challenge, we derived an efficient algorithm for making a Kronecker-factored approximation to $\b{\Lambda}_k = \b{F}_k + \b{\Lambda}_{k-1} \approx \b{A}_k \otimes \b{G}_k$ when $\b{\Lambda}_{k-1} = \b{A}_{k-1} \otimes \b{G}_{k-1}$ and $\b{F}_k$ are also Kronecker products.
This approximation minimizes the KL-divergence between $\mathcal{N}(\b{\mu}_k, (\b{A}_k \otimes \b{G}_k)^{-1})$ and $\mathcal{N}(\b{\mu}_k, (\b{\Lambda}_{k-1} + \b{F}_{k})^{-1})$ (see~\CrefNoLink{sec:kron_sums} for details).
Before training on the first task, we assume a spherical Gaussian prior $\b{\theta} \sim \mathcal{N}(\b{0}, p_{w}^{-2} \b{I})$.
The scale parameter $p_w$ can either be set to a fixed value (e.g. 1) or treated as a hyperparameter, and we optimize $p_w$ explicitly for our experiments in feedforward networks.
NCL also has a parameter $\alpha$ which is used to stabilize the matrix inversion $\b{\Lambda}_{k-1}^{-1} \approx (\b{A}_{k-1} \otimes \b{G}_{k-1} + \alpha^2 \b{I})^{-1}$ (\CrefNoLink{sec:implementation}).
This is equivalent to a hyperparameter used for such matrix inversions in OWM \citep{zeng2019continual} and DOWM \citep{duncker2020organizing}, and it is important for good performance with these methods.
The $p_w$ and $\alpha$ are largely redundant for NCL, and we generally prefer to fix $\alpha$ to a small value ($10^{-10}$) and optimize the $p_w$ only.
However, for our experiments in RNNs, we instead fix $p_w=1$ and perform a hyperparameter optimization over $\alpha$ for a more direct comparison with OWM and DOWM.
The NCL algorithm is described in pseudocode in \CrefNoLink{sec:implementation} together with additional implementation and computational details.
Finally, while we have derived NCL with a Laplace approximation in this section for simplicity, it can similarly be applied in the variational continual learning framework of \citet{nguyen2017variational} (\CrefNoLink{sec:vcl}).
Our code is available online\footnote{\href{https://github.com/tachukao/ncl}{https://github.com//tachukao/ncl}}.

\subsection{Related work}
\label{subsec:related}

As discussed in \Cref{subsec:bcl}, our method is derived from prior work that relies on Bayesian inference to perform weight regularization for continual learning \citep{kirkpatrick2017overcoming, nguyen2017variational,huszar2017quadratic,ritter2018online}.
However, we also take inspiration from the literature on natural gradient descent \citep{amari1998natural,kunstner2019limitations} to introduce a preconditioner that encourages parameter updates primarily in flat directions of previously learned tasks (\CrefNoLink{sec:ngd}).

Recent projection-based methods \citep{duncker2020organizing,zeng2019continual,saha2021gradient} have addressed the continual learning problem using an update rule of the form
\begin{equation}
	\label{eq:proj_up}
	\b{\theta} \leftarrow \b{\theta} + \gamma \b{P}_L \nabla_{\b{\theta}} \ell_k(\b{\theta}) \b{P}_R,
\end{equation}
where $\b{P}_L$ and $\b{P}_R$ are projection matrices constructed from previous tasks which encourage parameter updates that do not interfere with performance on these tasks.
Using Kronecker identities, we can rewrite \Cref{eq:proj_up} as
\begin{equation}
	\b{\theta} \leftarrow \b{\theta} + \gamma (\b{P}_R \otimes \b{P}_L) \nabla_{\b{\theta}} \ell_k(\b{\theta}).
\end{equation}
This resembles the NCL update rule in \Cref{eq:ncl_update} where we identify $\b{P}_R \otimes \b{P}_L$ with the approximate inverse prior precision matrix used for gradient preconditioning in NCL, $ \b{\Lambda}_{k-1}^{-1}= \b{A}_{k-1}^{-1} \otimes \b{G}_{k-1}^{-1}$.
Indeed, we note that for a Kronecker-structured approximation to $\b{F}_k$, the matrix $\b{A}_{k-1}$ approximates the empirical covariance matrix of the network activations experienced during all tasks up to $k-1$ (\citealp{martens2015optimizing,bernacchia2018exact}, \CrefNoLink{sec:fisher_deriv}), which is exactly the inverse of the projection matrix $\b{P}_R$ used in previous work \citep{duncker2020organizing,zeng2019continual}.
We thus see that NCL takes the form of recent projection-based continual learning algorithms with two notable differences:
\vspace{0.2em}\\
(i) NCL uses a \emph{left} projection matrix $\b{P}_L$ designed to approximate the posterior covariance of previous tasks $\b{\Lambda}_{k-1}^{-1} \approx \b{P}_R \otimes \b{P}_L$ (i.e., the prior covariance on task $k$; \CrefNoLink{sec:fisher_deriv}), while \citet{zeng2019continual} use the identity matrix $\b{I}$ and \citet{duncker2020organizing} use the covariance of recurrent inputs (\CrefNoLink{sec:proj}).
Notably, both of these choices of $\b{P}_L$ still provide reasonable approximations to $\b{\Lambda}_{k-1}^{-1}$, and thus the parameter updates of OWM and DOWM can also be viewed as projecting out steep directions of the prior on task $k$ (\CrefNoLink{sec:proj}).
\vspace{0.2em}\\
(ii) NCL includes an additional regularization term $(\b{\theta} - \b{\mu}_{k-1})$ derived from the Bayesian posterior objective, while \citet{duncker2020organizing} and \citet{zeng2019continual} do not use such regularization.
%
%
Importantly, this means that while NCL has a similar preconditioner and optimization path to these projection based methods, NCL has stationary points at the modes of the approximate Bayesian posterior while the stationary points of OWM and DOWM do not incorporate prior information from previous tasks (c.f.\ \Cref{fig:schematic}).

It is also interesting to note that previous Bayesian continual learning algorithms include a hyperparameter $\lambda$ that scales the prior compared to the likelihood term for the current task \citep{loo2020generalized}:
\begin{equation}
	\label{eq:L_lambda}
	\mathcal{L}_k^{(\lambda)}( \b{\theta} ) = \log p(\mathcal{D}_k | \b{\theta}) - \lambda (\b{\theta} - \b{\mu}_{k-1})^\top \b{\Lambda}_{k-1} (\b{\theta} - \b{\mu}_{k-1}).
\end{equation}
To minimize this loss and thus find a mode of the approximate posterior, it is common to employ pseudo-second-order stochastic gradient-based optimization algorithms such as Adam~\citep{kingma2014adam} that use their own gradient preconditioner based on an approximation to the Hessian of \Cref{eq:L_lambda}.
Interestingly, this Hessian is given by $\b{H}_k = -H(\mathcal{D}_k, \b{\theta}) - \lambda \b{\Lambda}_{k-1}$, which in the limit of large $\lambda$ becomes increasingly similar to preconditioning with the prior precision as in NCL.
Consistent with this, previous work using the online Laplace approximation has found that large values of $\lambda$ are generally required for good performance \citep{kirkpatrick2017overcoming, ritter2018online, van2018generative}.
Recent work has also combined Bayesian continual learning with natural gradient descent \citep{osawa2019practical, tseran2018natural}, and in this case a relatively high value of $\lambda=100$ was similarly found to maximize performance \citep{osawa2019practical}.

\section{Experiments and results}
\label{sec:exps}

\subsection{NCL in feedforward networks}
\label{subsec:feedforward}

To verify the utility of NCL for continual learning, we first compared our algorithm to standard methods in feedforward networks across two continual learning benchmarks: split MNIST and split CIFAR-100~(see \CrefNoLink{sec:tasks} for task details).
For each benchmark, we considered three continual learning settings~\citep{van2019three}.
In the `task-incremental' setting, task identity is available to the network at test time, in our case via a multi-head output layer~\citep{chaudhry2018riemannian}.
In the `domain-incremental' setting, task identity is unavailable at test time, and the output layer is shared between all tasks.
Finally, in the `class-incremental' setting, the network has to both infer task identity and solve the task, in our case by performing classification over all possible classes irrespective of which task the input in question is drawn from.

\begin{figure}[!t]
    \centering
    \includegraphics[width = 0.99\textwidth, trim={0 0 0 0}, clip=true]{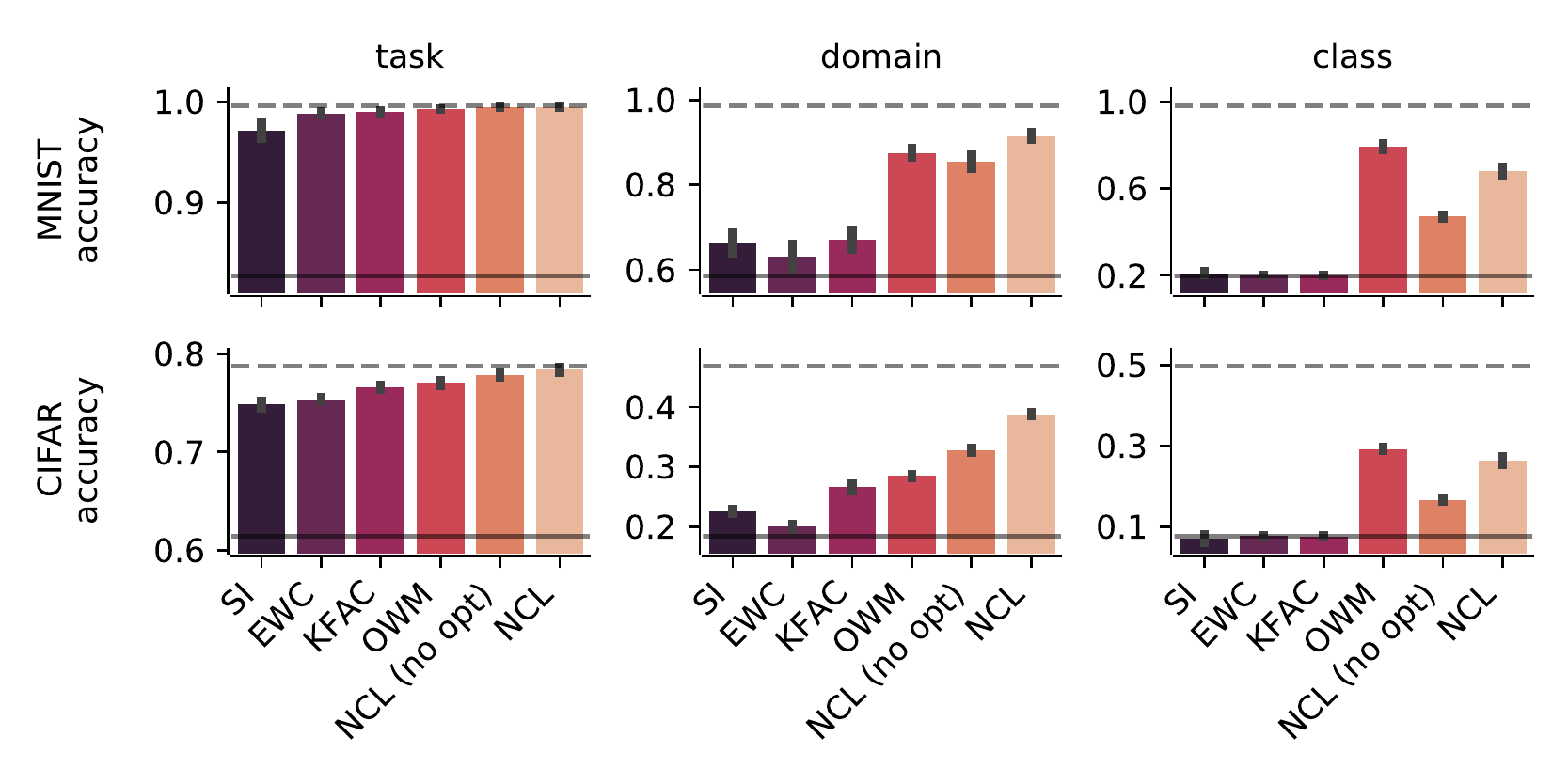}
    \vspace{-0.5em}
    \caption{\label{fig:ff}
        {\bfseries NCL performance in feedforward networks.}
        Average test accuracy after learning all tasks on split MNIST (top row) and split CIFAR-100 (bottom-row) in the task-, domain- and class-incremental learning settings.
        Dashed horizontal lines denote average performance when networks are trained simultaneously on all tasks. 
        Solid horizontal lines denote average performance when networks are trained sequentially on each task without applying any continual learning methods.
        Error bars denote standard error across $20$ (MNIST) or $10$ (CIFAR) random seeds.
        `NCL' indicates natural continual learning where the initial prior has been optimized on a held-out random seed, and `NCL (no-prior)' indicates NCL with a simple unit Gaussian prior and no hyperparameter optimization.
        Numerical results for these experiments are provided in \Cref{tab:numerical_results} in \Cref{sec:numerical_results}.
    }
    \vspace*{-1em}
\end{figure}

\citeauthor{van2019three} previously showed that parameter regularization methods such as EWC perform poorly in the domain- and class-incremental settings~\citep{van2019three}.
We therefore applied NCL as well as synaptic intelligence [SI; \citealp{zenke2017continual}], online EWC \citep{schwarz2018progress}, Kronecker factored EWC [KFAC; \citealp{ritter2018online}], and orthogonal weight modification [OWM; \citealp{zeng2019continual}] to split MNIST and split CIFAR-100 in the task-, domain- and class-incremental learning settings.
For these continual learning problems, we found that NCL outperformed all the baseline methods in the task- and domain-incremental learning settings (\Cref{fig:ff}).
In the class-incremental settings, we found that NCL performed comparably to but slightly worse than OWM.
However, both OWM and NCL comfortably outperformed the other compared methods in this setting.
These results suggest that the subpar performance of parameter regularization methods can be alleviated by regularizing their optimization paths, particularly in the domain- and class-incremental learning settings.

For the split MNIST and split CIFAR-100 experiments, each baseline method had a single hyperparameter ($c$ for SI, $\lambda$ for EWC and KFAC, $\alpha$ for OWM, and $p_w$ for NCL; \CrefNoLink{sec:implementation}) that was optimized on a held-out seed (see \Cref{sec:hp_opt}).
However, by setting the NCL prior to a unit Gaussian, we were also able to achieve good performance across task sets in a hyperparameter-free setting, further highlighting the robustness of the method (see ``NCL (no opt)'' in \Cref{fig:ff}).

\subsection{NCL in recurrent neural networks}
\label{subsec:ngcl_rnn}
We then proceeded to consider how NCL compares to previous methods in recurrent neural networks (RNNs), a setting that has recently proven challenging for continual learning~\citep{duncker2020organizing,ehret2020continual} and which is of interest to the study of continual learning in biological circuits \citep{duncker2020organizing,yang2019task}.
In these experiments, the task identity is available to the RNN (i.e., we consider the task-incremental learning setting).

\paragraph{Stimulus-response tasks}
\label{subsec:ryang}

\begin{figure}[!t]
    \centering
    \includegraphics[width = 0.95\textwidth, trim={0 0 0 0}, clip=true]{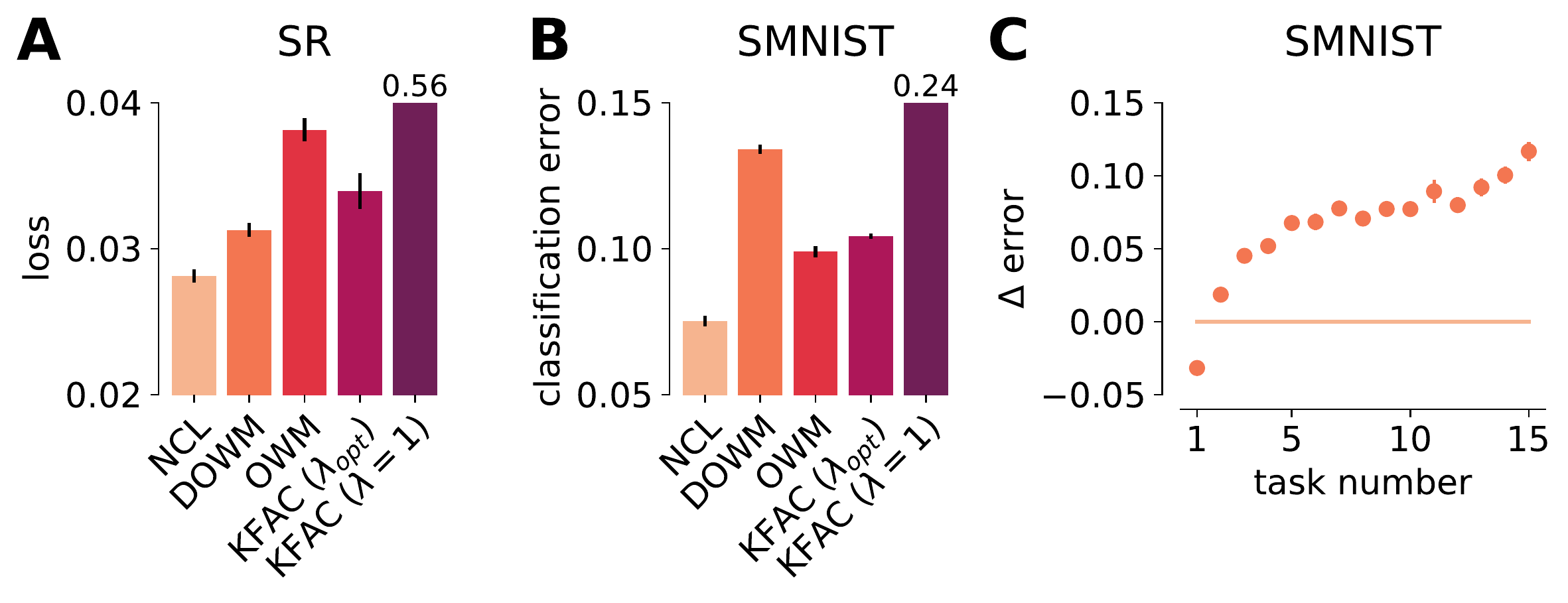}
    \caption{\label{fig:low_cap}
        {\bfseries Performance on SR and SMNIST tasks.}
        {\bfseries (A)}~Mean loss of NCL, DOWM, OWM, KFAC (optimal $\lambda$), and KFAC ($\lambda=1$) across stimulus-response tasks after sequential training on all tasks.
        Error bars indicate standard error across 5 random seeds.
        Here and in (B), KFAC with $\lambda=1$ failed catastrophically, and its performance is indicated in text as it does not fit on the axes.
        {\bfseries (B)}~Mean classification error across SMNIST tasks after sequential training.
        {\bfseries (C)}~Difference between the mean classification error of Laplace-DOWM and NCL as a function of task number.
        Error bars in (B) and (C) indicate standard error across 100 random task permutations.
        }
    \vspace*{-1em}
\end{figure}

In this section, we consider a set of neuroscience inspired `stimulus-response' (SR) tasks~(\citealp{yang2019task}; details in \CrefNoLink{sec:tasks}).
We first compared the performance and behavior of NCL to OWM, the top performing method in the feedforward setting (\Cref{fig:ff}), and to the projection-based DOWM method designed explicitly for RNNs \citep{duncker2020organizing}.
For a more direct comparison with OWM and DOWM, we fixed the NCL prior to a unit Gaussian for all RNN experiments and instead performed a hyperparameter optimization over `$\alpha$' used to regularize the matrix inversions for all three methods (\Cref{subsec:ngcl}, \CrefNoLink{sec:implementation}, \CrefNoLink{sec:hp_opt}, \citealp{duncker2020organizing}, \citealp{zeng2019continual}).
Following previous work, we trained RNNs with 256 recurrent units to sequentially solve six stimulus-response tasks~\citep{yang2009stably,duncker2020organizing}.
While NCL, OWM and DOWM all managed to learn the six tasks without catastrophic forgetting, we found that NCL achieved superior average performance across tasks after training (\Cref{fig:low_cap}A).

We then compared NCL, OWM, and DOWM to KFAC, the top performing parameter regularization method in our feedforward experiments (\Cref{fig:ff}) which uses Adam~\citep{kingma2014adam} to optimize the objective in \Cref{eq:laplace_update_mu} with a Kronecker-factored approximation to the posterior precision matrix (\Cref{subsec:bcl}; \citealp{ritter2018online}).
Consistent with the results shown in \citet{duncker2020organizing}, we found that NCL, OWM, and DOWM outperformed KFAC with $\lambda = 1$~(\Cref{fig:low_cap}A; see also \citealp{duncker2020organizing} for a comparison of DOWM and EWC).
We note that NCL and KFAC optimize the same objective function (\Cref{eq:laplace_update_mu}) and approximate the posterior precision matrix in the same way, but they differ in the way they precondition the gradient of the objective.
These results thus demonstrate empirically that the choice of optimization algorithm is important to prevent forgetting, consistent with the intuition provided by \Cref{fig:schematic}.

In feedforward networks, poor performance with weight regularization approaches such as EWC and KFAC has been mitigated by optimizing the hyperparameter $\lambda$, which increases the importance of the prior term compared to a standard Bayesian treatment~(\Cref{eq:L_lambda}; \Cref{subsec:feedforward}, \citealp{kirkpatrick2017overcoming,ritter2018online,loo2020generalized}).
We confirmed this here by performing a grid search over $\lambda$, which showed that KFAC with $\lambda \in [100, 1000]$ could perform comparably to the projection-based methods (\CrefNoLink{sec:lambda}; \Cref{fig:low_cap}A).
We hypothesize that the good performance provided by high $\lambda$ is partly due to the approximate second order nature of Adam which, together with the relative increase in the prior term compared to the data term, leads to preconditioning with a matrix resembling the prior $\b{\Lambda}_{k-1}$ (\Cref{subsec:related}).
In support of this hypothesis, we found that the KL divergence between the Adam preconditioner and the approximate prior precision $\b{\Lambda}_{k-1}$ decreased with increasing $\lambda$, and that the performance of KFAC with Adam could also be rescued by increasing $\lambda$ only when computing the preconditioner while retaining $\lambda=1$ when computing the gradients (\CrefNoLink{sec:lambda}).

\paragraph{Stroke MNIST}
\label{subsec:smnist}

One way to challenge the continual learning algorithms further is to increase the number of tasks.
We thus considered an augmented version of the stroke MNIST dataset [SMNIST; \citealp{de2016incremental}].
The original dataset consists of the MNIST digits transformed into pen strokes with the direction of the stroke at each time point provided as an input to the network.
Similar to \citet{ehret2020continual}, we constructed a continual learning problem by considering consecutive binary classification tasks inspired by the split MNIST task set.
We further increased the number of tasks by including a set of extra digits where the x and y dimensions have been swapped in the input stroke data, and another set where both the x and y dimensions have changed sign.
We also added high-variance noise to the inputs to increase the task difficulty.
This gave rise to a total of 15 binary classification tasks, each with unique digits not used in other tasks, which we sought to learn in a continual fashion using an RNN with 30 recurrent units (see \CrefNoLink{sec:tasks} for details).

As for the SR task set in \Cref{subsec:ryang}, we found that NCL outperformed previous projection-based methods (\Cref{fig:low_cap}B).
We again found that weight regularization with a KFAC approximation performed poorly with $\lambda = 1$, and that this poor performance could be partially rescued by optimizing over $\lambda$ (\Cref{fig:low_cap}B).
To investigate how the difference in performance between NCL and DOWM was affected by their different approximations to the Fisher matrix (\CrefNoLink{sec:proj}), we implemented NCL using the DOWM projection matrices as an alternative approximation to the inverse Fisher matrix.
We refer to this method as Laplace-DOWM.
We then considered how the performance on each task at the end of training depended on task number, averaged over different task permutations (\Cref{fig:low_cap}C).
We found that while Laplace-DOWM outperformed NCL on the first task, this method generally performed worse on subsequent tasks.
Notably, Laplace-DOWM exhibited a near-monotonic decrease in relative performance with task number, which is consistent with the intuition that DOWM overestimates the dimensionality of the parameter subspace that matters for previous tasks (\CrefNoLink{sec:proj}).
In contrast, although neural circuits are known to use orthogonal subspaces in different contexts, there is no general sense that learning more tasks in the past should systematically hinder learning in future contexts for biological agents.

\subsection{Dissecting the dynamics of networks trained on the SMNIST task set}
\label{subsec:dynamics}

\begin{figure}[!t]
    \centering
    \includegraphics[width = 0.98\textwidth, trim={0 0 0 0}, clip=true]{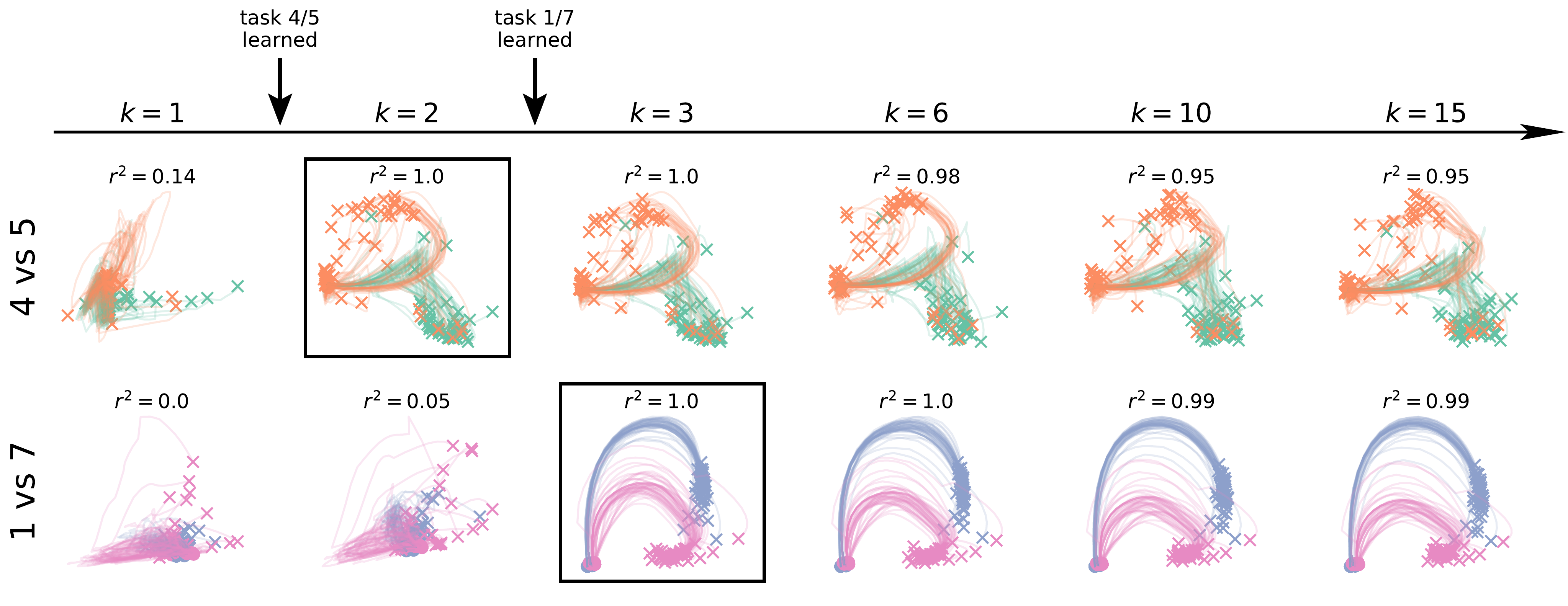}
    \caption{\label{fig:smnist_dyn}
        {\bfseries Latent dynamics during SMNIST.}
        We considered two example tasks, 4 vs 5 (top) and 1 vs 7 (bottom).
        For each task, we simulated the response of a network trained by NCL to 100 digits drawn from that task distribution at different times during learning.
        We then fitted a factor analysis model for each example task to the response of the network right after the correponding task had been learned (squares; $k = 2$ and $k=3$ respectively).
        We used this model to project the responses at different times during learning into a common latent space for each example task.
        For both example tasks, the network initially exhibited variable dynamics with no clear separation of inputs and subsequently acquired stable dynamics after learning to solve the task.
        The $r^2$ values above each plot indicate the similarity of neural population activity with that collected immediately after learning the corresponding task, quantified across all neurons (not just the 2D projection).
        }
    \vspace*{-1em}
\end{figure}

To further investigate how the trained RNNs solve the continual learning problems and how this relates to the neuroscience literature, we dissected the dynamics of networks trained on the SMNIST task set using the NCL algorithm.
To do this, we analyzed latent representations of the RNN activity trajectories, as is commonly done to study the collective dynamics of artificial and biological networks \citep{yu2009gaussian, gallego2020long, jensen2020manifold, mante2013context,jensen2021scalable}.
We considered two consecutive classification tasks, namely classifying 4's vs 5's ($k=2$) and classifying 1's vs 7's ($k=3$).
For each of these tasks, we trained a factor analysis model right after the task was learned, using network activity collected while presenting 50 examples of each of the two input digits associated with the task.
We then tracked the network responses to the same set of stimuli at various stages of learning, both before and after the task in question was acquired, using the trained factor analysis model to visualize low-dimensional summaries of the dynamics (\Cref{fig:smnist_dyn}).

Consistent with the network having successfully learned to solve these two tasks, we found that latent trajectories diverged over time for the two types of inputs in each task.
Critically, these diverging dynamics only emerged after the task was learned, and remained highly stable thereafter (\Cref{fig:smnist_dyn}).
The stability of the task-associated representations is consistent with recent work in the neuroscience literature showing that, in a primate reaching task, latent neural trajectories remain stable after learning \citep{gallego2020long}.
Since here we have access to the activity of all neurons throughout the task, we proceeded to quantify the source of this stability at the level of single units.
The stability of such single-neuron dynamics after learning has recently been a topic of much interest in biological circuits \citep{clopath2017variance, lutcke2013steady, rule2019causes}.
In the RNNs, we found that the single-unit representations of a given digit changed during learning of the task involving that digit but stabilized after learning, consistent with work in several distinct biological circuits \citep{peters2014emergence,katlowitz2018stable,dhawale2017automated,chestek2007single,ganguly2009emergence,jensen2021long}.
Similar results were found using the DOWM algorithm, which was explicitly designed to preserve network dynamics on previously learned tasks~\citep{duncker2020organizing}. Interestingly, the stable task representations learned by NCL and DOWM differed markedly from a network trained with replay for continual learning, which instead led to task representations that continued to change after initial task acquisition (\CrefNoLink{sec:dynamics_dowm}).
This illustrates how different approaches to continual learning can lead to qualitatively different circuit dynamics, and it suggests the use of continual learning in artificial networks as a model system for biological continual learning.

\section{Discussion}
\label{sec:discussion}
In summary, we have developed a new framework for continual learning based on approximate Bayesian inference combined with trust-region optimization.
We showed that this framework encompasses recent projection-based methods and found that it performs better than naive weight regularization.
This was particularly evident when task identity was not provided at test time and in recurrent neural networks, settings which have previously been challenging for many continual learning algorithms \citep{duncker2020organizing,ehret2020continual,van2019three}.
Furthermore, we showed that our principled probabilistic approach outperforms previous projection-based methods \citep{duncker2020organizing, zeng2019continual}, in particular when the number of tasks and their complexity challenges the network's capacity.
Finally, we analyzed the dynamics of the learned RNNs in a sequential binary classification problem, where we found that the latent dynamics adapt to each new task.
We also found that the task-associated dynamics were subsequently conserved during further learning, consistent with experimental reports of stable neural representations \citep{dhawale2017automated, gallego2020long, jensen2021long}.
Importantly, our results suggest that preconditioning with the prior covariance can lead to improved performance over existing continual learning algorithms.
In future work, it will therefore be interesting to apply this idea to other weight regularization approaches such as EWC with a diagonal approximate posterior \citep{kirkpatrick2017overcoming}.
Finally, a separate branch of continual learning utilizes replay-like mechanisms to reduce catastrophic forgetting~\citep{van2018generative,pan2020continual,li2017learning,shin2017continual,cong2020gan,titsias2020functional}.
While our work has focused on weight regularization, such regularization and replay are not mutually exclusive.
Instead, these two approaches have been found to further improve robustness to catastrophic forgetting when combined~\citep{nguyen2017variational,van2020brain}.

\paragraph{Impact and limitations}
While we have shown that NCL represents an important conceptual and methodological advance for continual learning, it also comes with several limitations.
One such limitation arises from the relative difficulty of computing the prior Fisher matrix which is needed for our projection step.
Indeed the success of methods such as Adam \citep{kingma2014adam} and EWC \citep{kirkpatrick2017overcoming} is due in part to their ease of implementation which facilitates broad applicability.
It will therefore be interesting to investigate how approximations such as a running average of a diagonal approximation to the empirical Fisher matrix as used in Adam could facilitate the development of simple yet powerful variants of NCL.

Furthermore, while NCL mitigates the need to overcount the prior from previous tasks via $\lambda$ as in KFAC, it does introduce two other (largely redundant) hyperparameters in the form of (i) the scale of the prior before the first task, and (ii) the parameter $\alpha$ used to regularize the inversion of the prior Fisher matrix, similar to OWM and DOWM \citep{duncker2020organizing, zeng2019continual}.
While $\alpha$ is an important hyperparameter for OWM and DOWM and we also optimize it in the RNN setting for a more direct comparison (\Cref{subsec:ngcl_rnn}), we find it more natural to set this parameter to a constant small value present only for numerical stability (\CrefNoLink{sec:implementation}).
This leaves the prior scale which we optimize explicitly in the feedforward setting (\Cref{subsec:feedforward}).
However, in future work it would be interesting to consider whether a good prior can be determined in a data free manner to make NCL a hyperparameter-free method.
Finally, computing the Fisher matrix used for pre-conditioning requires explicit knowledge of task boundaries.
In future work, it will therefore be interesting to develop an algorithm similar to NCL which also works for online learning problems with continually changing task distributions.

Addressing these challenges is important since machine learning algorithms increasingly need to be robust to changing data distributions and dynamic task specifications as they become more prominent in our everyday lives.
Much work has therefore gone into the development of methods for continual learning in the machine learning community.
However, with the increasing prevalence of practical algorithms for continual learning, it also becomes increasingly important that we understand how and why these algorithms work -- insights that can also help us understand when they might fail.
In this work, we have therefore attempted to shed light on the relationship between recent methods for continual learning as well as developing a new algorithm with a principled probabilistic interpretation that makes its underlying assumptions more explicit.
Taken together, we hope that this work will help improve our understanding of methods for continual learning while also providing an avenue for further research to increase the reliability and robustness of future continual learning algorithms.

\section*{Acknowledgements}
We are grateful to Siddharth Swaroop, Lea Duncker, Laura Driscoll, Naama Kadmon Harpaz, and Yashar Ahmadian for insightful discussions.
We thank Siddharth Swaroop and Robert Pinsler for useful comments on the manuscript. 

\bibliographystyle{apalike}
\bibliography{references}

\clearpage

\begin{appendices}
\crefalias{section}{appendix}

\section{Derivation of the NCL learning rule}
\label{sec:alt_ncl}
In this section, we provide further details of how the NCL learning rule in \CrefNoLink{subsec:ngcl} is derived and also provide an alternative derivation of the algorithm.

\paragraph{NCL learning rule}
As discussed in \CrefNoLink{subsec:ngcl}, we derive NCL as the solution of a trust region optimization problem.
That is, we maximize the posterior loss $\mathcal{L}_k(\b{\theta})$ within a region of radius $r$ centered around $\b{\theta}$ with a distance metric of the form $d(\b{\theta}, \b{\theta}+\b{\delta}) = \sqrt{\b{\delta}^\top \b{\Lambda}_{k-1} \b{\delta}/2}$.
This distance metric was chosen to take into account the curvature of the prior via its precision matrix $\b{\Lambda}_{k-1}$ and encourage parameter updates that do not affect performance on previous tasks.
Formally, we solve the optimization problem
\begin{align}
	\label{eq:ncl_opt}
	  & \b{\delta} 
	=
	\argmin_{\b{\delta}}
	\mathcal{L}_k(\b{\theta})
	+
	\nabla_{\b{\theta}} \mathcal{L}_k(\b{\theta})^\top
	\b{\delta}
	\quad
	\text{subject to}
	\;\;
	\frac{1}{2}\b{\delta}^\top \b{\Lambda}_{k-1} \b{\delta} \leq r^2,
\end{align}
where
$
\mathcal{L}_k(\b{\theta} + \b{\delta}) \approx 
\mathcal{L}_k(\b{\theta}) +
\nabla_{\b{\theta}} \mathcal{L}_k(\b{\theta})^\top \b{\delta}$
is a first-order approximation to the updated Laplace objective.
Here we recall from \CrefNoLink{eq:laplace_update_mu} that
\begin{equation}
	\mathcal{L}_k(\b{\theta}) = \ell_k(\b{\theta}) 
	- \frac12 (\b{\theta} - \b{\mu}_{k-1})^T \b{\Lambda}_{k-1} (\b{\theta} - \b{\mu}_{k-1})
\end{equation}
from which we get
\begin{equation}
	\nabla_{\b{\theta}} \mathcal{L}_k(\b{\theta})^\top \b{\delta}
	= \nabla_{\b{\theta}} \ell_k(\b{\theta})^\top \b{\delta} - (\b{\theta} - \b{\mu}_{k-1})^\top \b{\Lambda}_{k-1} \b{\delta}
\end{equation}
The optimization in \Cref{eq:ncl_opt} is carried out by introducing a Lagrange multiplier $\eta$ to construct a Lagrangian $\tilde{\mathcal{L}}$:
\begin{equation}
	\tilde{\mathcal{L}}(\b{\delta}, \eta) = 
	\mathcal{L}_k(\b{\theta})
	+
	\nabla_{\b{\theta}} \ell_k(\b{\theta})^\top \b{\delta}
	- (\b{\theta} - \b{\mu}_{k-1})^\top \b{\Lambda}_{k-1} \b{\delta}
	+ \eta (r^2 - \frac{1}{2}\b{\delta}^\top \b{\Lambda}_{k-1} \b{\delta}).
\end{equation}
We then take the derivative of $\tilde{\mathcal{L}}$ w.r.t. $\b{\delta}$ and set it to zero:
\begin{equation}
	\nabla_{\b{\delta}} \tilde{\mathcal{L}}(\b{\delta}, \eta)
	= \nabla_{\b{\theta}} \ell_k(\b{\theta})
	- \b{\Lambda}_{k-1} (\b{\theta} - \b{\mu}_{k-1})
	- \eta \b{\Lambda}_{k-1} \b{\delta}'
	= 0.
\end{equation}
Rearranging this equation gives
\begin{equation}
	\b{\delta} = \frac{1}{\eta} \left [
		\b{\Lambda}_{k-1}^{-1} \nabla_{\b{\theta}} \ell_k(\b{\theta})
		- (\b{\theta} - \b{\mu}_{k-1}),
	\right ].
\end{equation}
where $\eta$ itself depends on $r^2$ implicitly.
Finally we define a learning rate parameter $\gamma = 1/\eta$ and arrive at the NCL learning rule:
\begin{equation}
	\b{\theta} \leftarrow \b{\theta}
	+\gamma
	\left [
		\b{\Lambda}_{k-1}^{-1} \nabla_{\b{\theta}}
		\ell_k(\b{\theta})
		-
		(\b\theta - \b{\mu}_{k-1})
	\right ].
\end{equation}

\paragraph{Alternative derivation}

Here, we present an alternative derivation of the NCL learning rule. 
In this formulation, we seek to update the parameters of our model on task $k$ by maximizing $\mathcal{L}_k(\b{\theta})$ subject to a constraint on the allowed change in the prior term.
To find our parameter updates $\b{\delta}$, we again solve a constrained optimization problem:
\begin{align}
	\b{\delta} 
	  & = 
	\argmin_{\b{\delta}}
	\mathcal{L}_k(\b{\theta})
	+
	\nabla_{\b{\theta}} \mathcal{L}_k(\b{\theta})^\top
	\b{\delta}
	\quad
	\text{such that} \quad
	\;
	\mathcal{C}(\b{\delta}) 
	\leq r^2.
\end{align}
Here we define $\mathcal{C}(\b{\delta})$ as the approximate change in log probability under the prior
\begin{equation}
	\mathcal{C}(\b{\delta}) 
	=    
	(\b{\theta} + \b{\delta} - \b{\mu}_{k-1})^\top
	\b{\Lambda}_{k-1}
	(\b{\theta} + \b{\delta} - \b{\mu}_{k-1})
	-
	(\b{\theta} - \b{\mu}_{k-1})^\top
	\b{\Lambda}_{k-1}
	(\b{\theta} - \b{\mu}_{k-1}).
\end{equation}

Following a similar derivation to above, we find the solution to this optimization problem as
\begin{align}
	\eta \b{\delta} =                                                
	\b{\Lambda}_{k-1}^{-1}           
	\nabla_{\b{\theta}} \mathcal{L}_k(\b{\theta})                                    
	-                                     
	\eta (\b{\theta} - \b{\mu}_{k-1})    
	= \b{\Lambda}_{k-1}^{-1}  \nabla_{\b{\theta}} \ell_k(\b{\theta}) - (1+\eta) (\b{\theta} - \b{\mu}_{k-1}) 
\end{align}
for some Lagrange multiplier $\eta$.
This gives rise to the update rule
\begin{equation}
	\b{\theta} \leftarrow \b{\theta} 
	+\gamma 
	\left [ 
		\b{\Lambda}_{k-1}^{-1} \nabla_{\b{\theta}} 
		\ell_k(\b{\theta})
		- 
		\lambda
		(\b{\theta} - \b{\mu}_{k-1})
	\right ] 
\end{equation}
for a learning rate parameter $\gamma$ and some choice of the parameter $\lambda$ that depends on both $\eta$ and $\gamma$.
We recover the learning rule derived in \CrefNoLink{subsec:ngcl} with the choice of $\lambda=1$.
In practice, $\lambda$ can also be treated as a hyperparameter to be optimized (\Cref{sec:lambda}).

\section{Task details}
\label{sec:tasks}
\paragraph{Split MNIST}
	The split MNIST benchmark involves 5 tasks, each corresponding to the pairwise classification of two digits. The 10 digits of the MNIST dataset are randomly divided over the 5 tasks (i.e., for each random seed, this division can be different). During the incremental training protocol, these tasks are visited one after the other, followed by testing on all tasks.
	The original $28\times28$ pixel grey-scale images and the standard train/test-split are used, giving 60,000 training ($\sim$6,000 per digit) and 10,000 test images ($\sim$1,000 per digit).

\paragraph{Split CIFAR-100}
	The split CIFAR-100 benchmark consists of 10 tasks, with each task corresponding to a ten-way classification problem. The 100 classes of the CIFAR-100 dataset are randomly divided over the 10 tasks. Each network is trained on these tasks one after the other followed by testing on all tasks.
	The $32\times32$ pixel RGB-colour images are normalised by z-scoring each channel (using means and standard deviations calculated over the training set). We use the standard train/test-split, giving 500 training and 100 test images for each class.

\paragraph{Stimulus-response tasks} 
Here, we provide a brief overview of the six stimulus-response (SR) tasks.
Detailed descriptions of the stimulus-response tasks used in this work can be found in the appendix of \citetAPP{yang2019task}.
All tasks are characterized by a stimulus period and a response period, and some tasks include an additional delay period between the two.
The duration of the stimulus and delay periods are variable across trials and drawn uniformly at random within an allowed range.
During the stimulus period, the input to the network takes the form of $\b{x} = (\cos \theta_{in}, \sin \theta_{in} )$, where $\theta_{in} \in [0, 2\pi]$ is some stimulus drawn uniformly at random for each trial.
An additional tonic input is provided to the network which indicates the identity of the task using a one-hot encoding.
A constant input to a `fixation channel' during the stimulus and delay periods signifies that the network output should be 0 in the response channels and 1 in a `fixation channel'.
During the response period, the fixation input is removed and the output should be 0 in the fixation channel.
The target output in the response channels takes the form $\b{y} = (\cos \theta_{out}, \sin \theta_{out})$ where $\theta_{out}$ is some target output direction described for each task below:
\begin{itemize}
	\item \textbf{task 1 (fdgo)} During this task $\theta_{out} = \theta_{in}$ and there is no delay period.
	\item \textbf{task 2 (fdanti)} During this task $\theta_{out} = 2\pi - \theta_{in}$ and there is no delay period.
	\item \textbf{task 3 (delaygo)} During this task $\theta_{out} = \theta_{in}$ and there is a delay period separating the stimulus and response periods.
	\item \textbf{task 4 (delayanti)} During this task $\theta_{out} = 2\pi - \theta_{in}$ and there is a delay period separating the stimulus and response periods.
	\item \textbf{task 5 (dm1)} During this task, two stimuli are drawn from $[0, 2\pi]$ with different input magnitudes such that $\b{x} = (m_1 \cos \theta_1 + m_2 \cos \theta_2, m_1 \sin \theta_1 + m_2 \sin \theta_2 )$.
	      $\theta_{out}$ is then the element in $(\theta_1, \theta_2)$ corresponding to the largest $m$.
	\item \textbf{task 6 (dm2)} As in `dm1', but where the input is now provided through a separate input channel.
\end{itemize}
The loss for each task was computed as a mean squared error from the target output.

\paragraph{SMNIST}
For this task set, we use the stroke MNIST dataset created by \citetAPP{de2016incrementalAPP}.
This consists of a series of digits, each of which is represented as a sequence of vectors $\{ \b{x}_t \in \mathbb{R}^4 \}$.
The first two columns take values in $[-1, 0, 1]$ and indicate the discretized displacement in the x and y direction at each time step.
The last two columns are used for special `end-of-line' inputs when the virtual pen is lifted from the paper for a new stroke to start, and an `end-of-digit' input when the digit is finished.
See \citetAPP{de2016incrementalAPP} for further details about how the dataset was generated and formatted.
In addition to the standard digits 0-9, we include two additional sets of digits:
\begin{itemize}
	\item the digits 0-9 where the x and y directions have been swapped (i.e. the first two elements of $\b{x}_t$ are swapped),
	\item the digits 0-9 where the x and y directions have been inverted (i.e. the first two elements of $\b{x}_t$ are negated).
\end{itemize}
Furthermore, we omitted the initial entry of each digit corresponding to the `start' location to increase task difficulty.
We turned this dataset into a continual learning task by constructing five binary classification tasks for each set of digits: $\{[2,3], [4,5], [1,7], [8,9], [0,6] \}$.
Note that we have swapped the `1' and `6' from a standard split MNIST task to avoid including the 0 vs 1 classification task which we found to be too easy.
For each trial, a digit was sampled at random from the corresponding dataset, and $\b{x}_t$ was provided as an input to the network at each time step corrupted by Gaussian noise with $\sigma = 1$.
After the `end-of-digit' input, a response period with a duration of 5 time steps followed.
During this response period only, a cross-entropy loss was applied to the output units $\b{y}$ to train the network.
During testing, digits were sampled from the separate test dataset and classification performance was quantified as the fraction of digits for which the correct class was assigned the highest probability in the last timestep of the response period.
Task identity was provided to the network, which was used in the form of a multi-head output layer.

\section{Network architectures}
\label{sec:network_archictecture}
\paragraph{Feedforward network archictecture}

For split MNIST, all methods are compared using a fully-connected network with 2 hidden layers containing 400 units with ReLU non-linearities, followed by a softmax output layer.

For split CIFAR-100, the network consists of 5 pre-trained convolutional layers, 2 fully-connected layers with 2000 ReLU units each and a softmax output layer. The architecture of the convolutional layers and their pre-training protocol on the CIFAR-10 dataset are described in~\citep{van2020brain}. The only difference is that here we pre-train a new set of convolutional layers for each random seed, while in~\citep{van2020brain} the same set of pre-trained convolutional layers was used for all random seeds. For all compared methods, the pre-trained convolutional layers are frozen during the incremental training protocol.

The softmax output layer of the feedforward networks is treated differently depending on the continual learning setting~\citep{van2019three}. In the task-incremental learning setting, there is a separate output layer for each task and only the output layer of the task under consideration is used at any given time (i.e., a multi-head output layer). In the domain-incremental learning setting, there is a single output layer that is shared between all tasks. In the class-incremental learning setting, there is one large output layer that spans all tasks and contains a separate output unit for each class.

%
%

\paragraph{Recurrent network architecture}
\label{subsec:network_achitecture_rnn}

The dynamics of the RNN used in \Cref{subsec:ngcl_rnn} can be described by the following equations:
\begin{align}
	\label{eq:dynamics}
	\b{h}_t & = \b{H} \b{r}_{t-1} +  \b{G}\b{x}_t
	+ \b{\xi}_t = \b{W} \b{z}_t + \b{\xi}_t       \\
	\b{y}_t & \sim p(\b{y}_t | \b{C} \b{r}_t)
\end{align}
where we define $\b{r}_t = \phi(\b{h}_t)$, $\b{z}_t = (\b{r}_{t-1}^\top, \b{x}_t^\top)^\top$, $\b{W} = (\b{H}^\top, \b{G}^\top)^\top$, and time is indexed by $t$.
Here, $\b{r} \in \mathbb{R}^{N_{rec} \times 1}$ are the network activations, $\b{x} \in \mathbb{R}^{n_{in} \times 1}$ are the inputs, $\b{y} \in \mathbb{R}^{n_{out} \times 1}$ are the network outputs, and we refer to $\b{W} \b{z}_t$ as the `recurrent inputs' to the network.
The noise model $p(\b{y}_t | \b{C}\b{r}_t)$ may be a Gaussian distribution for a regression task or a categorical distribution for a classification task, and $\phi(\b{h})$ is a nonlinearity that is applied to $\b{h}$ element-wise (in this work the ReLU function).
The parameters of the RNN are given by $\b{\theta} = (\b{W}, \b{C})$.
The process noise $\{ \b{\xi}_t \}$ are zero-mean Gaussian random variables with covariance matrices $\b{\Sigma}^{\b{\xi}}_t$.
In this model, the log-likelihood of observing a sequence of outputs $\b{y}_1, \ldots, \b{y}_T$ given inputs $\b{x}_1, \ldots, \b{x}_T$ and $\b{\xi}_1, \ldots, \b{\xi}_T$ is given by
\begin{equation}
	\mathcal{\ell}(\b{\theta}) =
	\log p_\theta(\{ \b{y} \} | \{ \b{x} \}, \{ \b{\xi} \})
	=
	\log p( \{\b{y} \} | \{ \b{C} \b{r} \})
	\label{eq:ll},
\end{equation}
where $p(\b{y} | \b{C} \b{r})$ may be a Gaussian distribution for a regression task or a categorical distribution for a classification task.

\section{KFAC approximation to the Fisher matrix}
\label{sec:fisher_deriv}
For all experiments in this work, we make a Kronecker-factored approximation to the FIM of each task $k$ in \Cref{eq:fisher_k}.
Concretely, we use the block-wise Kronecker-factored approximation to the FIM proposed in Section 3 of \citetAPP{martens2015optimizingAPP} for feedforward neural networks.
For recurrent neural networks, we use the approximation presented in Section 3.4 of \citetAPP{martens2018kroneckerAPP}.
Both approximations allow us to write the FIM on task $k$ as the Kronecker product $\b{F}_{k} \approx \hat{\b{A}}_{k} \otimes \hat{\b{G}}_k$.
For completeness, we derive the approximation for RNNs below.
We refer the readers to \citetAPP{martens2015optimizingAPP} for details on derivations for feedforward networks.


\paragraph{KFAC approximation for RNNs}
Recall from \Cref{subsec:network_achitecture_rnn} that the log likelihood of observing a sequence of outputs $\b{y}_1, \ldots, \b{y}_T$ given inputs $\b{x}_1, \ldots, \b{x}_T$ and $\b{\xi}_1, \ldots, \b{\xi}_T$ is
\begin{equation}
    \ell(\b{W}, \b{C})
    =  \sum_{t=1}^T \log p(\b{y}_t | \b{C} \b{r}_t),
\end{equation}
where $\b{r}_t$ is completely determined by the dynamics of the network and the inputs.
With a slight abuse of notation, we use $\overline{\b{x}}$ to denote both $\partial \ell/\partial \b{x}$ for vectors $\b{x}$ and $\partial \ell/\partial \text{vec}(\b{X})$ for matrices $\b{X}$.
In this section, it should be clear given the context whether $\overline{\b{x}}$ is representing the gradient of $\mathcal{L}$ with respect to a vector or a vectorized matrix.
Using these notations, we can write the gradient of $\mathcal{L}$ with respect to $\text{vec}(\b{W})$ as :
\begin{align}
    \overline{\b{w}}
    =
    \sum_{t=1}^T
    \overline{\b{h}}_t
    \frac{\partial \b{h}_t}{\partial \text{vec}(\b{W})}
    =
    \sum_{t=1}^T
    \overline{\b{h}}_t
    \b{z}_t^\top
    =
    \sum_{t=1}^T
    \b{z}_t \otimes \overline{\b{h}}_t
\end{align}
which can be easily derived fom the backpropagation through time (BPTT) algorithm and the definition of a Kronecker product.
Using this expression for $\overline{\b{w}}$, we can write the FIM of $\b{W}$ as:
\begin{align}
    \b{F}_{\b{W}}
     & =
    \mathbb{E}_{
    \{ (\b{\xi}, \b{x}, \b{y}) \} \sim \mathcal{M}
    }
    \left [
        \overline{\b{w}}
        \,
        \overline{\b{w}}^\top
        \right ] \\
     & =
    \mathbb{E}
    \left [
        \left (
        \sum_{t=1}^T
        \b{z}_t \otimes \overline{\b{h}}_t
        \right )
        \left (
        \sum_{s=1}^T
        \b{z}_s \otimes \overline{\b{h}}_s
        \right )^{\top}
        \right ] \\
     & =
    \sum_{t=1}^T
    \sum_{s=1}^T
    \mathbb{E}
    \left [
        \left (
        \b{z}_t
        \b{z}_s^\top
        \right )
        \otimes
        \left (
        \overline{\b{h}}_t
        \overline{\b{h}}_s^\top
        \right )
        \right ]                           .
\end{align}
Here the expectations are taken with respect to the model distribution.
Unfortunately, computing $\b{F}_{\b{W}}$ can be prohibitively expensive.
First, the number of computations scales quadratically with the length of the input sequence $T$.
Second, for networks of dimension $n$, there are $n^4$ entries in the Fisher matrix which can therefore be too large to store in memory, let alone perform any useful computations with it.
For this reason, we follow \citetAPP{martens2018kroneckerAPP} and make the following three assumptions in order to derive a tractable Kronecker-factored approximation to the Fisher.
The first assumption we make is that the input and recurrent activty $\b{z}_t$ is uncorrelated with the adjoint activations $\overline{\b{h}}_t$:
\begin{align}
    \b{F}_{\b{W}}
    \approx
    \sum_{t=1}^T
    \sum_{s=1}^T
    \mathbb{E}
    \left [
        \b{z}_t
        \b{z}_s^\top
        \right ]
    \otimes
    \mathbb{E}_{
    \{ (\b{\xi}, \b{x}, \b{y}) \} \sim \mathcal{M}
    }
    \left [
        \overline{\b{h}}_t
        \overline{\b{h}}_s^\top
        \right ].
\end{align}
Note that this approximation is exact when the network dynamics are linear (i.e., $\phi(\b{x}) = \b{x}$).
The second assumption that we make is that both the forward activity $\b{z}_t$ and adjoint activity $\overline{\b{h}}_t$ are temporally homogeneous.
That is, the statistical relationship between $\b{z}_t$ and $\b{z}_s$ only depends on the difference $\tau = s-t$, and similarly for that between $\overline{\b{h}}_t$ and $\overline{\b{h}}_s$.
Defining
$\mathcal{A}_\tau = \mathbb{E} \left [ \b{z}_s \b{z}_{s+\tau}^\top \right ]$
and similarly
$\mathcal{G}_\tau = \mathbb{E} \left [ \overline{\b{h}}_s \overline{\b{h}}_{s+\tau}^\top \right ]$, we have $\mathcal{A}_{-\tau} = \mathcal{A}_\tau^\top$ and $\mathcal{G}_{-\tau} = \mathcal{G}_\tau$.
Using these expressions, we can further approximate the Fisher as:
\begin{align}
    \b{F}_{\b{W}}
     &
    \approx
    \sum_{\tau=-T}^T
    (T - |\tau|)
    \mathcal{A}_\tau
    \otimes
    \mathcal{G}_\tau.
\end{align}
The third and final approximation we make is that $\mathcal{A}_\tau \approx 0$ and $\mathcal{G}_\tau \approx 0$ for $\tau \neq 0$.
In other words, we assume the forward activity $\b{z}_t$ and adjoint activity $\overline{\b{h}}_t$ are approximately indendent across time.
This gives the final expression:
\begin{equation}
    \b{F}_{\b{W}} \approx
    \mathbb{E} \left [ T \right ]
    \,
    \mathbb{E}
    \left [ \b{z} \b{z}^\top \right ] \otimes
    \mathbb{E}
    \left [ \overline{\b{h}} \, \overline{\b{h}}^\top \right ]
    = \b{\hat{A}}_{\b{W}} \otimes \b{\hat{G}}_{\b{W}},
\end{equation}
where we have also taken an expectation over the sequence length $T$ to account for variable sequence lengths in the data.
Following a similar derivation, we can approximate the Fisher of $\b{C}$ as:
\begin{equation}
    \b{F}_{\b{C}}
    \approx
    \mathbb{E} \left [ T  \right ]
    \, \mathbb{E}
    \left [
        \b{r} \b{r}^\top
        \right ]
    \otimes
    \mathbb{E}
    \left [
        \overline{\b{y}} \, \overline{\b{y}}^\top
        \right ]
    = \b{\hat{A}}_{\b{C}} \otimes \b{\hat{G}}_{\b{C}}.
\end{equation}

The quality of these assumptions and comparisons with the `approximate Fisher matrices' used in OWM and DOWM are discussed in \CrefNoLink{sec:proj}.

\section{Implementation}
\label{sec:implementation}
\begin{algorithm}[H]
	\DontPrintSemicolon
	\SetKw{Parameters}{parameters: }
	\SetKw{Input}{input: }
	\SetKw{Initialize}{initialize: }
	\Input $f$ (network), $\{ \mathcal{D}_k \}_{k=1}^K$, $\alpha$, $p_w$ (prior), $B$ (batch size), $\gamma$ (learning rate), $\theta_0$, $\rho$
	\;
	\Initialize 
	$\b{A}_{\theta} \leftarrow p_w \b{I}$,  
	$\b{G}_{\theta} \leftarrow p_w \b{I}$,
	\;
	\Initialize 
	$\theta_1 \leftarrow \theta_0$,
	\Initialize 
	$\b{M}_{\theta} \leftarrow \text{zeros\_like}(\theta_0)$,
	\tcp*{Gradient momentum}
	\For{$k=1 \ldots K$}{

		$\widetilde{\b{A}}, \widetilde{\b{G}} \leftarrow \text{nearest\_kf\_sum}(\b{A}_\theta \otimes \b{G}_\theta, \alpha \b{I} \otimes \alpha \b{I})$\tcp*{\Cref{sec:kron_sums}}
		$\b{P}_L \leftarrow \widetilde{\b{G}}^{-1}$\;
		$\b{P}_R \leftarrow \widetilde{\b{A}}^{-1}$

		\While{not converged}{
			$\{\b{x}^{(i)}, \b{y}^{(i)} \}_{i=1}^{B}\sim \mathcal{D}_k$ \tcp*{Input and target output}

			\For{$i=1, \ldots, B$}{
				$\hat{\b{y}}^{(i)} = f(\b{x}^{(i)}, \b{\theta}_k)$ \tcp*{Empirical output}
			}
			$\ell = 
			\sum_i^B \log p(\b{y}^{(i)} | \hat{\b{y}}^{(i)})/B$
			\tcp*{Loss}
			\;
																		
			\% Build up momentum \;
			$\b{M}_{\theta} \leftarrow \rho \b{M}_{\theta} + \nabla_{\theta} \ell + \b{G}_{\theta} (\theta_k - \theta_{k-1}) \b{A}_{\theta}$\;
			\;
																					
			\% Update model parameters \;
			$\theta_k \leftarrow \theta_k - \gamma p_w^2\; 
			\b{P}_L \; \b{M}_{\theta} \b{P}_R
			$ 
		}
		\% Update Fisher matrix components \;
		Compute $\hat{\b{A}}_{k}$ and ${\hat{\b{G}}}_{k}$\tcp*{\Cref{sec:fisher_deriv}}
		$\b{A}_{\theta}, \b{G}_{\theta} \leftarrow 
		\text{nearest\_kf\_sum}(
		\b{A}_{\theta} \otimes \b{G}_{\theta},
		\hat{\b{A}}_{k} \otimes \hat{\b{G}}_{k})$\tcp*{\Cref{sec:kron_sums}}
	}
	\caption{NCL with momentum}
	\label{alg:ncl}
\end{algorithm}

In this section we discuss various implementation details for NCL. 
\Cref{alg:ncl} provides an overview of the algorithm in the form of pseudocode.
For numerical stability, we add $\alpha^2 \b{I}$ to the precision matrix $\b{\Lambda}_{k-1}$ before computing the projection matrices $\b{P}_L$ and $\b{P}_R$.
In general, we set the prior over the parameters $\b{\theta}$ when learning the first task as $p(\b{\theta}) = \mathcal{N}(\b{0}; p_w^{-2} \b{I})$.

\paragraph{Feedforward networks}
By default, we set $p_w^{-2}$ to be approximately the number of samples that the learner sees in each task, corresponding to a unit Gaussian prior before normalizing our precision matrices by the amount of data seen in each task (here, $p_w^{-2} = 12000$ for split MNIST and $p_w^{-2} = 5000$ for split CIFAR-100).
We also consider hyperparameter optimizations over $p_w^{-2}$ by trying different values on a log scale from $10^2$ to $10^{11}$ with a random seed not included during the evaluation (see \Cref{sec:hp_opt}). We use $\alpha=10^{-10}$ and $\lambda=1$ for all experiments.

For all experiments with feedforward networks, we use a batch size of 256 and we train for either 2000 iterations per task (split MNIST) or 5000 iterations per task (split CIFAR-100).
For NCL and OWM, we train with momentum ($\rho = 0.9$) and a learning rate of $\gamma = 0.05$. For SI, EWC and KFAC, we train using the Adam optimizer ($\beta_1=0.9$, $\beta_2=0.999$) with learning rate of $\gamma=0.001$ (split MNIST) or $\gamma=0.0001$ (split CIFAR-100). All models were trained on single GPUs with training times of 10-100 minutes.

\paragraph{RNNs}
We again set $p_w^{-2}$ approximately equal to the number of samples that the learner sees in each task, corresponding to a unit Gaussian prior before normalizing our precision matrices by the amount of data seen in each task (here, $p_w^{-2} = 10^{6}$ for the stimulus-response task and $p_w^{-2} = 6000$ for SMNIST).

We used momentum ($\rho=0.9$) in all our experiments involving NCL, OWM and DOWM, as is also done in \citetAPP{duncker2020organizingAPP}.
We found that the use of momentum greatly speeds up convergence in practice.

All models were trained on single GPUs with training times of 10-100 minutes depending on the task set and model size.
We used a training batch size of 32 for the stimulus-response tasks and 256 for the SMNIST tasks.
In all cases, we used a test batch size of 2048 for evaluation and for computing projection and Fisher matrices.
We used a learning rate of $\gamma=0.01$ for SMNIST and $\gamma=0.005$ for the stimulus-response tasks across all projection-based methods.
We used a learning rate of $\gamma=0.001$ for KFAC with the Adam optimizer.
All models were trained on $10^6$ data samples per task.
A hyperparameter optimization over $\alpha$ for the projection-based methods and $\lambda$ for KFAC with Adam is provided in \Cref{sec:hp_opt}.

\section{Relation to projection-based continual learning}
\label{sec:proj}

\begin{figure}[!t]
    \centering
    \includegraphics[width = 0.99\textwidth, trim={0 0 0 0}, clip=true]{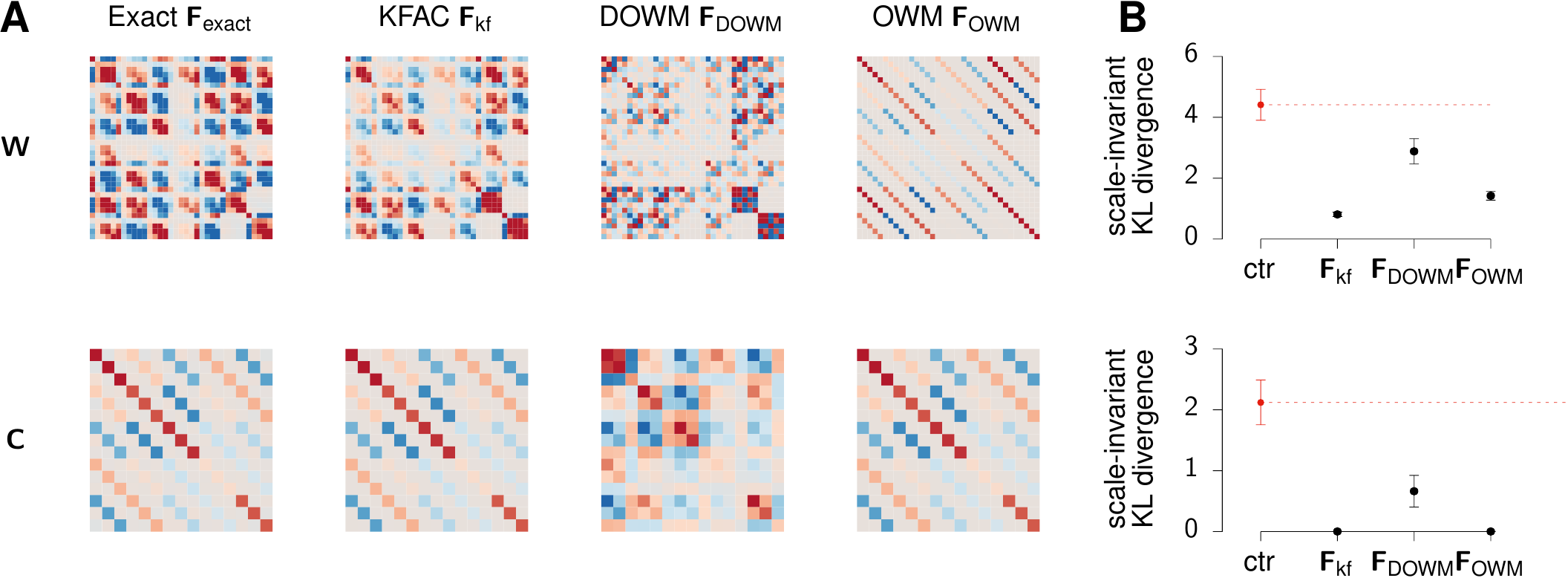}
    \caption{\label{fig:fisher_approx}
        \textbf{Comparison of projection matrices.}
        In a Bayesian framework, we can formalize what is meant by directions `important for previous tasks' as those that are strongly constrained by the prior $p(\b{\theta} | \mathcal{D}_{1:k-1})$.
        To see how this compares with OWM and DOWM, we considered the Kronecker-structured precision matrices $\b{F}_{\text{approx}}$ implied by the projection matrices $\b{P}_R$ and $\b{P}_L$ for each method and related them to the exact Fisher matrix $\b{F}_{\text{exact}}$ in a linear recurrent network.
        \textbf{(A; top)}~$\b{F}_{\text{exact}}$ (left) for $\b{W}$ as well as the approximations to $\b{F}_{\text{exact}}$ provided by our Kronecker-factored approximation (KFAC; $\b{F}_{\text{kf}}$), DOWM ($\b{F}_{\text{DOWM}}$), and OWM ($\b{F}_{\text{OWM}}$).
        \textbf{(B; top)}~Scale-invariant KL-divergence (\Cref{eq:scaled_kl}) between $\mathcal{N}(\b{\mu}, \b{F}_{\text{exact}}^{-1})$ and $\mathcal{N}(\b{\mu}, \b{F}_{\text{approx}}^{-1})$ for each approximation.
        Red horizontal line indicates the mean value obtained from $\b{F}_{\text{approx}} = \b{R} \b{F}_{\text{exact}} \b{R}^\top$ where $\b{R}$ is a random rotation matrix (averaged over 500 random samples).
        \textbf{(Bottom)}~Same as (A--B) but for the readout matrix $\b{C}$.
        }
    \vspace*{-1em}
\end{figure}

In this section, we further elaborate on the intuition that projection-based continual learning methods such as Orthogonal Weight Modification (OWM; \citealpAPP{zeng2019continualAPP}) may be viewed as variants of NCL with particular approximations to the prior Fisher matrix.
These approaches are typically motivated as a way to restrict parameter changes in a neural network that is learning a new task to subspaces orthogonal to those used in previous tasks. 

For example, to solve the continual learning problem in RNNs as described in \CrefNoLink{subsec:network_achitecture_rnn}, \citetAPP{duncker2020organizingAPP} proposed a projected gradient algorithm (DOWM) that restricts modifications to the recurrent/input weight matrix $\b{W}$ on task $k+1$ to column and row spaces of $\b{W}$ that are not heavily ``used'' in the first $k$ tasks.  
Specifically, they concatenate input and recurrent activity $\b{z}_t$ across the first $k$ tasks into a matrix $\b{Z}_{1:k}$. 
They use $\b{Z}_{1:k}$ and $\b{W}\b{Z}_{1:k}$ as estimates of the row and column spaces of $\b{W}$ that are important for the first $k$ tasks.
They proceed to construct the following projection matrices: 
\begin{align}
	\b{P}_z^{1:k}    
	  & =       
	\b{Z}_{1:k}
	(\b{Z}_{1:k} \b{Z}_{1:k}^\top + \alpha \b{I})^{-1}                  
	{\b{Z}_{1:k}}^\top\\
	  & \approx 
	k \alpha 
	\left (
	\mathbb{E} 
	\left [ \b{z}\b{z}^\top  
	\right ]
	+ \alpha \b{I} 
	\right )
	^{-1} \\
	\b{P}_{wz}^{1:k} 
	  & =       
	\b{W}\b{Z}_{1:k}
	(\b{W} \b{Z}_{1:k} \b{Z}_{1:k}^\top \b{W}^\top + \alpha \b{I})^{-1} 
	(\b{W}\b{Z}_{1:k})^\top\\
	  & \approx 
	k \alpha 
	\left (
	\b{W}
	\mathbb{E} 
	\left [ 
	\b{z}\b{z}^\top  
	\right ]
	\b{W}^\top 
	+ \alpha \b{I}
	\right)^{-1},
\end{align}
which are used to derive update rules for $\b{W}$ as:
\begin{align}
	\text{vec}(\Delta \b{W})
	  & \propto     
	\left (\b{P}_{z}^{1:k} \otimes \b{P}_{wz}^{1:k} \right ) 
	\overline{\b{w}}\\
	  & \propto 
	\left (
	\mathbb{E} 
	\left [ 
	\b{z}\b{z}^\top  
	\right ]
	+ \alpha \b{I}
	\right )^{-1}
	\otimes 
	\left (
	\b{W}
	\mathbb{E} 
	\left [ 
	\b{z}\b{z}^\top  
	\right ]
	\b{W}^\top 
	+ \alpha \b{I}
	\right ) ^{-1}
	\overline{\b{w}}
\end{align}
where $\overline{\b{w}} = \text{vec}(\nabla_{\b{W}} \ell_{k+1}(\b{W}, \b{C}))$.
These projection matrices restrict changes in the row and column space of $\b{W}$ to be orthogonal to $\b{Z}_{1:k}$ and $\b{W} \b{Z}_{1:k}$ respectively.
Similar update rules can be defined for $\b{C}$.
\citetAPP{zeng2019continualAPP} propose a similar projection-based learning rule (OWM) in feedforward networks, which only restricts changes in the row-space of the weight parameters (i.e., $\b{P}_{wz} = \b{I}$).

\begin{figure}[!t]
    \centering
    \includegraphics[width = 0.99\textwidth, trim={0 0 0 0}, clip=true]{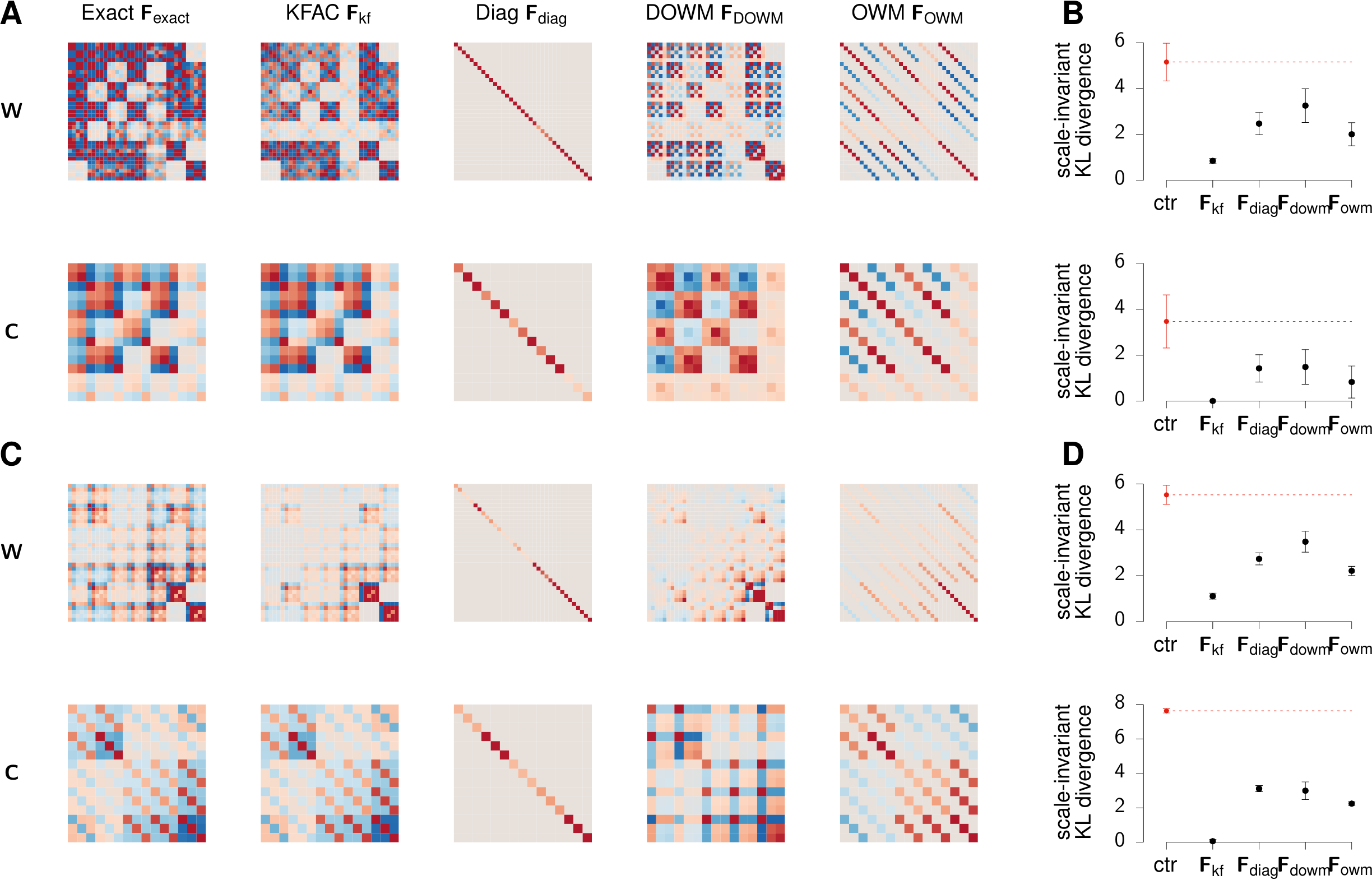}
    \caption{\label{fig:fisher_approx_supp}
        {\bfseries Comparison of Fisher Approximations in a Linear RNN with rotated Gaussian and categorical likelihoods.}
        {\bfseries (A)}~Exact and approximations to the Fisher information matrix of the recurrent and input weight matrix $\b{W}$ (left) and the linear readout $\b{C}$ (bottom) of a linear recurrent neural network with Gaussian noise and non-diagonal noise covariance $\b{\Sigma}$.
        From the left: exact Fisher information matrix $\b{F}_{\text{exact}}$, Kronecker-Factored approximation ($\b{F}_{\text{kf}}$; KFAC),  Diagonal ($\b{F}_{\text{diag}}$), DOWM ($\b{F}_{\text{DOWM}}$), and OWM ($\b{F}_{\text{OWM}}$).
        {\bfseries (B)}~Scale-invariant KL-divergence between $\mathcal{N}(\b{0}, \b{F}_{\text{exact}}^{-1})$ and $\mathcal{N}(\b{0}, \b{F}^{-1})$ for $\b{F} \in \{\b{F}_{\text{kf}},\b{F}_{\text{diag}}, \b{F}_{\text{DOWM}}, \b{F}_{\text{OWM}} \}$.
        Red horizontal lines indicate the mean value obtained from $\b{F}_{\text{approx}} = \b{R} \b{F}_{\text{exact}} \b{R}^\top$ where $\b{R}$ is a random rotation matrix (averaged over 500 random samples).
        {\bfseries (C-D)}~As in (A-B), now for a categorical noise model $p( \b{y} | \b{C} \b{r}) = \text{Cat} \left ( \text{softmax}(\b{C} \b{r}) \right )$.
        }
    \vspace*{-1em}
\end{figure}

With a scaled additive approximation to the sum of Kronecker products (see \Cref{sec:kron_sums}), the NCL update rule on task $k+1$  is given by
\begin{align}
	\text{vec}(\Delta \b{W})
	  & \propto 
	\left (
	\mathbb{E} 
	\left [ 
	\b{z}\b{z}^\top  
	\right ]
	+ \pi \alpha \b{I}
	\right )^{-1}
	\otimes 
	\left (
	\mathbb{E} 
	\left [ 
	\overline{\b{h}}
	\,
	\overline{\b{h}}^\top  
	\right ]
	+ \frac{1}{\pi} \alpha \b{I}
	\right ) ^{-1}
	\overline{\b{w}}
	+ (\vect (\b{W}_{k}) - \vect (\b{W})).
\end{align}
We see that this NCL update rule looks similar to the OWM and DOWM update steps, and that they share the same projection matrix in the row-space $\b{P}_z$ when $\pi = 1$.
The methods proposed by \citetAPP{duncker2020organizingAPP} and \citetAPP{zeng2019continualAPP} can thus be seen as approximations to NCL with a Kronecker structured Fisher matrix.
However, we also note that OWM and DOWM do not include the regularization term $(\vect (\b{W}_{k}) - \vect (\b{W}))$.
This implies that while OWM and DOWM encourage parameter updates along flat directions of the prior, the performance of these methods may deteriorate in the limit of infinite training duration if a local minimum of task $k$ is not found in a flat subspace of previous tasks (c.f. \CrefNoLink{fig:schematic}).

To further emphasize the relationship between OWM, DOWM and NCL, we compared the approximations to the Fisher matrix $\b{F}_{\text{approx}} = \b{P}_R^{-1} \otimes \b{P}_L^{-1}$ implied by the projection matrices of these methods (\CrefNoLink{fig:fisher_approx}).
Here we found that OWM and DOWM provided reasonable approximations to the true Fisher matrix with both Gaussian (\Cref{fig:fisher_approx}) and categorical (\Cref{fig:fisher_approx_supp}) observation models.
This motivates a Bayesian interpretation of these methods as using an approximate prior precision matrix to project gradients, similar to the derivation of NCL in \Cref{sec:alt_ncl}.
Here it is also worth noting that while we use an optimal sum of Kronecker factors to update the prior precision after each task in NCL (\Cref{sec:kron_sums}), OWM and DOWM simply sum their Kronecker factors.
In the case of OWM, this is in fact an exact approximation to the sum of the Kronecker products since the right Kronecker factor is in this case a constant matrix $\b{I}$.
For DOWM, summing the individual Kronecker factors does not provide an optimal approximation to the sum of the Kronecker products, but our results in \Cref{sec:kron_sums} suggest that it is a fairly reasonable approximation up to a scale factor which can be absorbed into the learning rate.

Another recent projection-based approach to continual learning developed by \citetAPP{saha2021gradient} restricts parameter updates to occur in a subspace of the full parameter space deemed important for previous tasks.
This method, known as `Gradient projection memory' (GPM), is similar to OWM but with a hard cut-off separating `important' from `unimportant' directions of parameter space.
The important subspace is in this case determined by thresholding the singular values of the activity matrix $\b{Z}_k$.
GPM can thus be seen as a discretized version of OWM with a projection matrix constituting a binary approximation to the prior Fisher matrix.

\section{Kronecker-factored approximation to the sums of Kronecker Products}
\label{sec:kron_sums}
In this section, we consider three different Kronecker-factored approximations to the sum of two Kronecker products:
\begin{equation}
	\b{X} \otimes \b{Y} \approx 
	\b{Z} = \b{A} \otimes \b{B} + \b{C} \otimes \b{D}.
\end{equation}
In particular, we consider the special case where 
$\b{A} \in \mathbb{R}^{n \times n}$, $\b{B} \in \mathbb{R}^{m \times m}$, $\b{C} \in \mathbb{R}^{n \times n}$, and $\b{D} \in \mathbb{R}^{m \times m}$ are symmetric positive-definite.
$\b{Z}$ will not in general be a Kronecker product, but for computational reasons it is desirable to approximate it as one to avoid computing or storing a full-sized precision matrix.

\paragraph{Scaled additive approximation}

The first approximation we consider was proposed by \citetAPP{martens2015optimizingAPP}. 
They propose to approximate the sum with 
\begin{equation}
	\b{Z}
	\approx 
	(\b{A} + \pi \b{C}) \otimes (\b{B} + \frac{1}{\pi} \b{D}),
\end{equation}
where $\pi$ is a scalar parameter.
Using the triangle inequality, \citetAPP{martens2015optimizingAPP} derived an upper-bound to the norm of the approximation error
\begin{align}
	  & \hphantom{=\,} 
	\|
	\b{Z} -
	(\b{A} + \pi \b{C}) \otimes (\b{B} + \frac{1}{\pi} \b{D})
	\|\\
	  & =              
	\|
	\frac{1}{\pi} \b{A} \otimes \b{D}
	+
	\pi 
	\b{C} \otimes \b{B} 
	\|\\
	  & \leq           
	\frac{1}{\pi} 
	\|
	\b{A} \otimes \b{D}
	\|
	+
	\pi 
	\|
	\b{C} \otimes \b{B} 
	\|
\end{align}
for any norm $\|\cdot \|$.
They then minimize this upper-bound with respect to $\pi$ to find the optimal $\pi$:
\begin{equation}
	\pi = \sqrt{ 
		\frac{\|\b{C} \otimes \b{B}\|}{\|\b{A} \otimes \b{D}\|} }.
\end{equation}
As in \citep{martens2015optimizing}, we use a trace norm in bounding the approximation error, and noting that $\text{Tr}(\b{X} \otimes \b{Y}) = \text{Tr}(\b{X}) \text{Tr}(\b{Y})$, we can compute the optimal $\pi$ as:
\begin{equation}
	\pi = \sqrt{ \frac{\text{Tr}(\b{B}) \text{Tr}(\b{C})}{ \text{Tr}(\b{A}) \text{Tr}(\b{D})}}.
\end{equation}

\paragraph{Minimal mean-squared error}
The second approximation we consider was originally proposed by \citetAPP{van1993approximationAPP}.
In this case, we approximate the sum of Kronecker products by minimizing a mean squared loss:
\begin{align}
	\b{X}, \b{Y} 
	  & = 
	\argmin_{\b{X}, \b{Y}} 
	\| 
	\b{Z}
	-\b{X} \otimes \b{Y}
	\|_{F}^2\\
	  & = 
	\argmin_{\b{X}, \b{Y}} 
	\| 
	\mathcal{R}(\b{A} \otimes \b{B})
	+\mathcal{R}(\b{C} \otimes \b{D})
	- \mathcal{R}(\b{X} \otimes \b{Y})
	\|_{F}^2\\
	  & = 
	\argmin_{\b{X}, \b{Y}} 
	\| 
	\text{vec}(\b{A})\text{vec}(\b{B})^\top
	+
	\text{vec}(\b{C})\text{vec}(\b{D})^\top
	-
	\text{vec}(\b{X})\text{vec}(\b{Y})^\top
	\|_{F}^2,
\end{align}
where $\mathcal{R}(\b{A} \otimes \b{B}) = \text{vec}(\b{A}) \text{vec}(\b{B})^\top$ is the rearrangement operator~\citepAPP{van1993approximationAPP}.
The optimization problem thus involves finding the best rank-one approximation to a rank-2 matrix. 
This can be solved efficiently using a singular value decomposition (SVD) without ever constructing an $n^2 \times m^2$ matrix (see \Cref{algo:kfac_sum_mse} for details).

\begin{algorithm}[H]
	\DontPrintSemicolon
	\SetKw{Parameters}{parameters: }
	\SetKw{Input}{input: }
	\SetKw{Initialize}{initialize: }
																
	\Input $\b{A}$, $\b{B}$, $\b{C}$, $\b{D}$
	\;
	$\b{a} \leftarrow \text{vec}(\b{A})$,
	$\b{b} \leftarrow \text{vec}(\b{B})$,
	$\b{c} \leftarrow \text{vec}(\b{C})$,
	$\b{d} \leftarrow \text{vec}(\b{D})$
	\tcp*{Vectorize $\b{A}, \b{B}, \b{C}, \b{D}$}
	$\b{Q}, \_ \leftarrow \text{QR}(\begin{bmatrix} \b{a} ; \b{c} \end{bmatrix})$ \tcp*{Orthogonal basis for $\b{a}$ and $\b{c}$ in $\mathbb{R}^{n^2 \times 2}$}
	$\b{H} \leftarrow 
	(\b{Q}^\top \b{a})\b{b}^\top
	+
	(\b{Q}^\top \b{c})\b{d}^\top
	$\;
	$\b{U},  \b{s}, \b{V}^\top \leftarrow \text{SVD}(\b{H})$\;
	$\b{y} \leftarrow \text{first column of } \sqrt{\b{s}_1}\b{V}$\;
	$\b{x} \leftarrow \text{first column of } \sqrt{\b{s}_1}\b{Q} \b{U}$\;
	$\b{X} \leftarrow \text{reshape}(\b{x}, (n, n))$,
	$\b{Y} \leftarrow \text{reshape}(\b{y}, (m, m))$
	\caption{Mean-squared error approximation of the sum of Kronecker products}
	\label{algo:kfac_sum_mse}
\end{algorithm}

\begin{figure}[!t]
    \centering
    \includegraphics[width = 0.85\textwidth, trim={0 0 0 0}, clip=true]{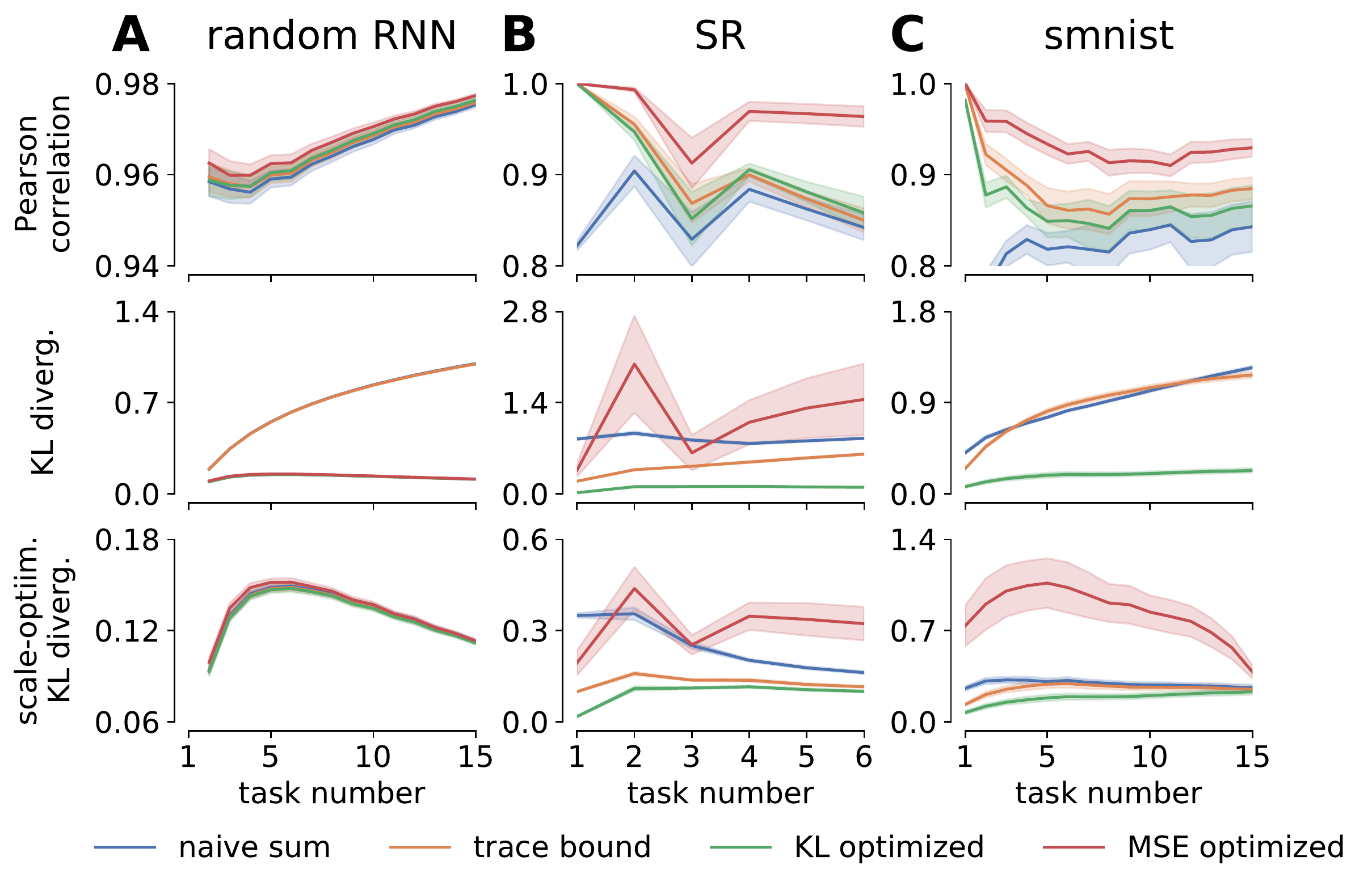}
    \caption{\label{fig:kf_sum}
        {\bfseries Comparison of different Kronecker approximations to consecutive sums of two Kronecker products.}
        {\bfseries (A)}~Comparison of approximations for Fisher matrices computed from random RNNs with dynamics as described in \CrefNoLink{subsec:ngcl_rnn}.
        All similarity/distance measures are computed between the true sum $\sum_{k'}^k \b{F}_{k'}$ and each iterative approximation.
        {\bfseries (B)}~As in (A) for the Fisher matrices from the stimulus-response tasks, here trained with 50 hidden units to make the computation of the true sum tractable.
        {\bfseries (C)}~As in (A) for the Fisher matrices from the SMNIST tasks.
        Note that the KL divergence for the MSE-minimizing approximation is not shown in panel 2 as it is an order of magnitude larger than the alternatives and thus does not fit on the axis.
        }
    \vspace*{-1em}
\end{figure}

\paragraph{Minimal KL-divergence}

In this paper, we propose an alternative approximation to $\b{Z}$ motivated by the fact that $\b{X} \otimes \b{Y}$ is meant to approximate the precision matrix of the approximate posterior after learning task $k$.
We thus define two multivariate Gaussian distributions $q(\b{w}) = \mathcal{N}(\b{w}; \b{\mu}, \b{X} \otimes \b{Y})$ and $p(\b{w}) =\mathcal{N}(\b{w}; \b{\mu}, \b{Z})$ (note that the mean of these distributions are found in NCL by gradient-based optimization).
We are interested in finding the matrices $\b{X}$ and $\b{Y}$ that minimize the KL-divergence between the two distributions
\begin{align}
	2D_\text{KL}(q || p)
	  & = 
	\log | \b{X} \otimes \b{Y} | - \log | \b{Z} | 
	+ \text{Tr}(\b{Z} (\b{X} \otimes \b{Y})^{-1})
	- d
	\\
	  & = 
	m \log | \b{X} | + n \log |\b{Y}|
	+ \text{Tr}(\b{A} \b{X}^{-1} \otimes \b{B} \b{Y}^{-1})
	+ \text{Tr}(\b{C}\b{X}^{-1} \otimes \b{D} \b{Y}^{-1})
	-d
	\\
	  & = 
	-m \log | \b{X}^{-1}| - n \log |\b{Y}^{-1}|
	+ \text{Tr}(\b{A}\b{X}^{-1}) \text{Tr}(\b{B} \b{Y}^{-1})\\
	  &   
	\qquad
	\qquad
	\qquad
	\qquad
	\qquad
	\qquad
	\qquad
	\qquad
	+ \text{Tr}(\b{C} \b{X}^{-1}) \text{Tr}(\b{D} \b{Y}^{-1})
	-d
\end{align}
where $d = nm$.
Differentiating with respect to $\b{X}^{-1}$, and $\b{Y}^{-1}$ and setting the result to zero, we get
\begin{align}
	0 & = 
	\frac{\partial D_{\text{KL}}(q||p)}{\partial \b{X}^{-1}}
	=
	\frac{1}{2}
	\left [
	-m \b{X}
	+ \text{Tr}(\b{B} \b{Y}^{-1}) \b{A}
	+ \text{Tr}(\b{D} \b{Y}^{-1}) \b{C}
	\right ]                     \\
	0 & = 
	\frac{\partial D_{\text{KL}}(q||p)}{\partial \b{Y}^{-1}}
	=
	\frac{1}{2}
	\left [
	-n \b{Y}
	+ \text{Tr}(\b{A} \b{X}^{-1}) \b{B}
	+ \text{Tr}(\b{C} \b{X}^{-1}) \b{D}
	\right ].
\end{align}
Rearranging these equations, we find the self-consistency equations:
\begin{align}
	\b{X}
	      & = 
	\frac{1}{m}
	\left [
	\text{Tr}(\b{B} \b{Y}^{-1}) \b{A}
	+ \text{Tr}(\b{D} \b{Y}^{-1}) \b{C}
	\right ] 
	\label{eq:X_consistency}
	\\
	\b{Y} & = 
	\frac{1}{n}
	\left [
	\text{Tr}(\b{A} \b{X}^{-1}) \b{B}
	+ \text{Tr}(\b{D} \b{X}^{-1}) \b{D}
	\right ]
	\label{eq:Y_consistency}
	.
\end{align}
This shows that the optimal $\b{X}$ ($\b{Y}$) is a linear combination of $\b{A}$ and $\b{C}$ ($\b{B}$ and $\b{D}$).
It is unclear whether we can solve  for $\b{X}$ and $\b{Y}$ analytically in \Cref{eq:X_consistency} and \Cref{eq:Y_consistency}.
However, we can find $\b{X}$ and $\b{Y}$ numerically by iteratively applying the following update rules:
\begin{align}
	\b{X}_{k+1} =                         
	(1-\beta)                             
	\b{X}_k                               
	+                                     
	\frac{\beta}m                         
	\left (                               
	\text{Tr}(\b{B} \b{Y}_k^{-1}) \b{A}   
	+ \text{Tr}(\b{D} \b{Y}_k^{-1}) \b{C} 
	\right )                              \\
	\b{Y}_{k+1} =                         
	(1-\beta)                             
	\b{Y}_k                               
	+                                     
	\frac{\beta}n                         
	\left (                               
	\text{Tr}(\b{A} \b{X}_k^{-1}) \b{C}   
	+ \text{Tr}(\b{C} \b{X}_k^{-1}) \b{D} 
	\right )                              
\end{align}
for initial guesses $\b{X}_0$ and $\b{Y}_0$.
In practice, we initialize using the scaled additive approximation and find that the algorithm converges with $\beta =0.3$ after tens of iterations.

\paragraph{Comparisons}
To compare different approximations of the precision matrix to the posterior, we consider Kronecker structured Fisher matrices from (i) a random RNN model, (ii) the Fishers learned in the stimulus-response tasks, and (iii) the Fishers learned in the SMNIST tasks.
We then iteratively update $\b{\Lambda}_k \approx \b{\Lambda}_{k-1} + \b{F}_k$, approximating this sum using each of the approaches described above as well as a naive unweighted sum of the pairs of Kronecker factors.
We compare these approximations using three different metrics: the correlation with the true sum of Kronecker products $\sum_{k'}^k \b{F}_{k'}$ (\Cref{fig:kf_sum}, top row), the KL divergence from the true sum (\Cref{fig:kf_sum}, middle row), and the scale-optimized KL divergence from the true sum (\Cref{fig:kf_sum}, bottom row).
Here we define the scale-optimized KL divergence as
\begin{align}
	\label{eq:scaled_kl}
	\text{KL}_\lambda[\b{\Lambda}_1 || \b{\Lambda}_2] 
	  & = \text{min}_\lambda \text{KL}[\lambda \b{\Lambda}_1 || \b{\Lambda}_2] \\
	  & =                                                                      
	\frac{1}2
	\left (
	\log 
	\frac{|\b{\Lambda}_1|}{|\b{\Lambda}_2|} + d \log \frac{\text{Tr}(\b{\Lambda}_1^{-1}\b{\Lambda}_2)}{d}
	\right ),
\end{align}
where $d$ is the dimensionality of the precision matrices $\b{\Lambda}_1$ and $\b{\Lambda}_2$ and we take $\text{KL}[\b{\Lambda}_1, \b{\Lambda}_2] = D_{\text{KL}}(\mathcal{N}(\b{0}, \b{\Lambda}_1^{-1}) || \mathcal{N}(\b{0}, \b{\Lambda}_2^{-1}))$.
This is a useful measure since a scaling of the approximate prior does not change the subspaces that are projected out in the weight projection methods but merely scales the learning rate.
By contrast in NCL, having an appropriate scaling is useful for a consistent Bayesian interpretation.

We find that all the methods yield reasonable correlations and scale-optimized KL divergences between the true sum of Kronecker products and the approximate sum, although the L2-optimized approximation tends to have a slightly better correlation and slightly worse scaled KL (\Cref{fig:kf_sum}, red).
However, the KL-optimized Kronecker sum greatly outperforms the other methods as quantified by the regular KL divergence and is the method used in this work since it is relatively cheap to compute and only needs to be computed once per task (\Cref{fig:kf_sum}, green).

\section{Natural gradient descent and the Fisher Information Matrix}
\label{sec:ngd}
When optimizing a model with stochastic gradient descent, the parameters $\b{\theta}$ are generally changed in the direction of steepest gradient of the loss function $\mathcal{L}$:
\begin{equation}
	\b{g} = \nabla_{\b{\theta}} \mathcal{L}.
\end{equation}
This gives rise to a learning rule
\begin{equation}
	\b{\theta}_{i+1} = \b{\theta}_i - \gamma \b{g}
\end{equation}
where $\gamma$ is a learning rate which is usually set to a small constant or updated according to some learning rate schedule.
However, we note that the parameter change itself has units of $ [ \b{\theta} ]^{-1} $ which suggests that such a naïve optimization procedure might be pathological under some circumstances.
Consider instead the more general definition of the normalized gradient $\hat{\b{g}}$:
\begin{equation}
	\hat{\b{g}} = \text{lim}_{\epsilon \rightarrow 0} \frac{1}{Z(\epsilon)} \text{argmin}_{\b{\delta}} \mathcal{L}(\b{\theta} + \b{\delta}) \hspace{2 cm} d(\b{\theta}, \b{\theta}+\b{\delta}) \leq \epsilon.
\end{equation}
Here, $\hat{\b{g}}$ is the direction in state space which minimizes $\mathcal{L}$ given a step of size $\epsilon$ according to some distance metric $d(\cdot, \cdot)$.
Canonical gradient descent is in this case recovered when $d(\cdot, \cdot)$ is Euclidean distance in parameter space 
\begin{equation}
	d(\b{\theta}, \b{\theta}') = ||\b{\theta} - \b{\theta}'||^2_2.
\end{equation}
We now formulate $\mathcal{L}(\b{\theta})$ as depending on a statistical model $p(\mathcal{D} | \b{\theta})$ such that $\mathcal{L}(\b{\theta}) = \mathcal{L}(p(\mathcal{D} | \b{\theta}))$.
This allows us to define the direction of steepest gradient in terms of the change in probability distributions
\begin{equation}
	d(\b{\theta}, \b{\theta}') = \text{KL} \left [ p(\mathcal{D} | \b{\theta}') || p(\mathcal{D} | \b{\theta}) \right ].
\end{equation}
It can be shown that the direction of steepest decent for small step sizes is in this case given by \citepAPP{kunstner2019limitationsAPP, amari1998naturalAPP}
\begin{equation}
	\b{g} \propto \b{F}^{-1} \nabla \mathcal{L}(\b{\theta}),
\end{equation}
where $\b{F}$ is the Fisher information matrix
\begin{equation}
	\b{F}(\b{\theta}) = \mathbb{E}_{p(\mathcal{D} | \b{\theta})} \left [ \nabla \log p(\mathcal{D} | \b{\theta}) \nabla \log p(\mathcal{D} | \b{\theta})^T \right ].
\end{equation}
We thus get an update rule of the form
\begin{equation}
	\b{\theta}_{i+1} = \b{\theta}_i - \gamma \b{F}^{-1} \nabla_{\b{\theta}} \mathcal{L},
\end{equation}
which has units of $[ \b{\theta} ]$ and corresponds to a step in the direction of parameter space that maximizes the decrease in $\mathcal{L}$ for an infinitesimal change in $p(\mathcal{D} | \b{\theta})$ as measured using KL divergences.
It has been shown in a large body of previous work that such natural gradient descent leads to improved performance \citepAPP{bernacchia2018exactAPP,osawa2019practicalAPP,amari1998naturalAPP}, and the main bottleneck to its implementation is usually the increased cost of computing $\b{F}$ or a suitable approximation to this quantity.

We note that this optimization method is very similar to that derived for NCL in \CrefNoLink{subsec:ngcl} and \Cref{sec:alt_ncl} except that NCL uses the approximate Fisher for \textit{previous} tasks instead of the Fisher information matrix of the current loss.
This is important since (i) it mitigates the need for computing a fairly expensive Fisher matrix at every update step, and (ii) it ensures that parameters are updated in directions that preserve the performance on previous tasks.

\section{Further results}
\subsection{Performance with different prior scalings}
\label{sec:lambda}

\begin{figure}[!t]
    \centering
    \includegraphics[width = 1.\textwidth]{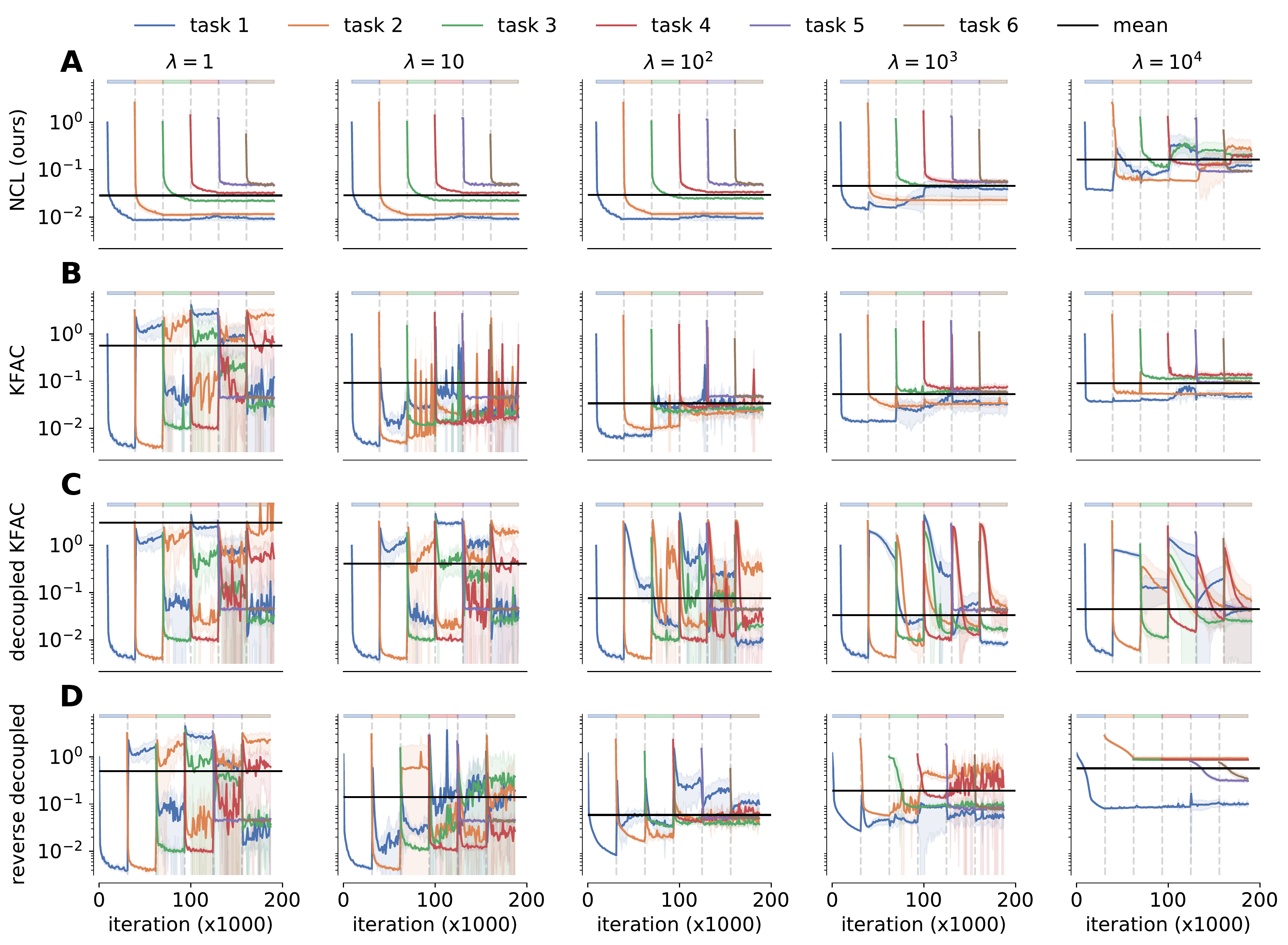}
    \caption{\label{fig:by_lambda}
        {\bfseries Continual learning on SR tasks with different $\lambda$ .}
        {\bfseries (A)}~Evolution of the loss during training for each of the six stimulus-response tasks for NCL with different values of $\lambda$.
        The performance of NCL is generally robust across different choices of $\lambda$ until it starts overfitting too heavily on early tasks.
        {\bfseries (B)}~As in (A), now for KFAC with Adam which performs poorly for small $\lambda$.
        {\bfseries (C)}~As in (B), now with ``decoupled Adam'' where we fix $\lambda_m=1$ for the gradient estimate and vary $\lambda=\lambda_v$ for the preconditioner (see \Cref{sec:lambda} for details).
        Interestingly, this is sufficient to overcome the catastrophic forgetting observed for KFAC with $\lambda_m = \lambda_v = 1$.
        The transient forgetting observed at the beginning of a new task is likely due to the time it takes to gradually update the preconditioner for the new task as more data is observed.
        {\bfseries (D)}~As in (C), now fixing $\lambda_v=1$ for the preconditioner and varying $\lambda=\lambda_m$ for the gradient estimate. 
        For higher values of $\lambda_m$, this performs worse than both KFAC and decoupled KFAC.
        }
    \vspace*{-1em}
\end{figure}

Here we consider the performance of KFAC and NCL for different values of $\lambda$ on the stimulus-response task set with 256 recurrent units.
We start by recalling that $\lambda$ is a parameter that is used to define a modified Laplace loss function with a rescaling of the prior term (c.f. \CrefNoLink{subsec:related}):
\begin{equation}
  \label{eq:L_lambda_app}
  \mathcal{L}_k^{(\lambda)}( \b{\theta} ) = \log p(\mathcal{D}_k | \b{\theta}) - \lambda (\b{\theta} - \b{\mu}_{k-1})^\top \b{\Lambda}_{k-1} (\b{\theta} - \b{\mu}_{k-1}).
\end{equation}
In this context, it is worth noting that KFAC and NCL have the same stationary points when they share the same value of $\lambda$.
Despite this, the performance of NCL was robust across different values of $\lambda$ (\Cref{fig:by_lambda}A), while learning was unstable and performance generally poor for KFAC with small values of $\lambda \in [1, 10]$.
However, as we increased $\lambda$ for KFAC, learning stabilized and catastrophic forgetting was mitigated (\Cref{fig:by_lambda}B).
A similar pattern was observed for the SMNIST task set (\Cref{sec:hp_opt}).

\begin{figure}[!t]
    \centering
    \includegraphics[width = 0.7\textwidth]{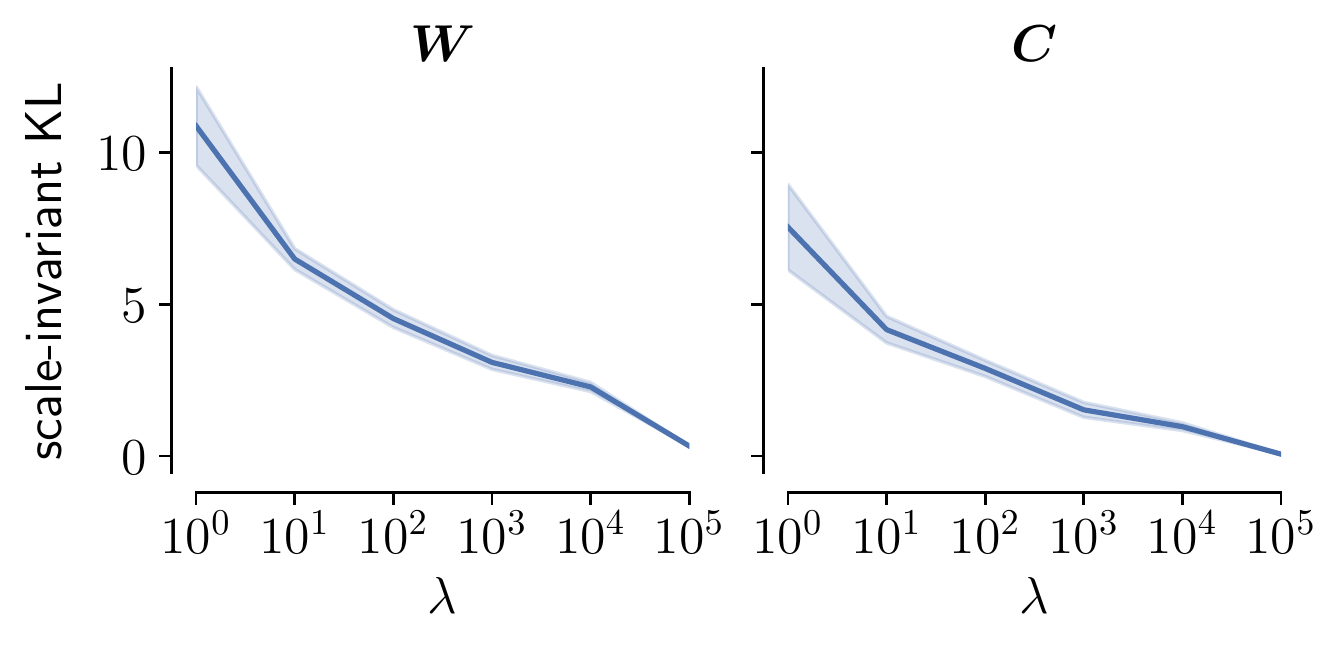}
    \caption{\label{fig:kfac_lambda_prec}
        {\bfseries Similarity of the Adam preconditioner and diagonal Fisher matrix.}
        Scale-invariant KL divergence (\Cref{eq:scaled_kl}) between the diagonal of $\b{\Lambda}_{k-1}$ and the preconditioner used by Adam ($\sqrt{\b{v}}$; \citealp{kingma2014adam}) at the end of training on task $k$.
        Results are averaged over the five first stimulus-response tasks, and the figure indicates mean and standard error across 5 seeds for the state matrix $\b{W}$ (left) and the output matrix $\b{C}$ (right).
        }
    \vspace*{-1em}
\end{figure}

We hypothesize that the improved performance of KFAC for high values of $\lambda$ is due in part to the gradient preconditioner of KFAC becoming increasingly similar to NCL's preconditioner $\b{\Lambda}_{k-1}^{-1}$ as $\lambda$ increases (\CrefNoLink{subsec:related}).
To test this hypothesis, we modified the Adam optimizer~\citepAPP{kingma2014adamAPP} to use different values of $\lambda$ when computing the Adam momentum and preconditioner.
Specifically, we computed the momentum and preconditioner of some scalar parameter $\theta$ as:
\begin{align}
  m^{(i)} & \leftarrow \beta_1 m^{(i-1)} + (1-\beta_1) \nabla_{\theta} \mathcal{L}^{(\lambda_m)}                    \\
  v^{(i)} & \leftarrow \beta_2 v^{(i-1)} + (1-\beta_2) \left ( \nabla_{\theta} \mathcal{L}^{(\lambda_v)} \right )^2
\end{align}
where $\mathcal{L}^{(\lambda)}$ is defined in \Cref{eq:L_lambda_app} and importantly $\lambda_m$ may not be equal to $\lambda_v$.
As in vanilla Adam, we used $m$ and $v$ to update the parameter $\theta$ according to the following update equations at the $i^{th}$ iteration:
\begin{align}
  \hat{m}^{(i)} & \leftarrow {m^{(i)}}/{(1 - \beta_1^i)} \\
  \hat{v}^{(i)} & \leftarrow {v^{(i)}}/{(1 - \beta_2^i)} \\
  \theta^{(i)}  & \leftarrow \theta^{(i-1)}
  + \gamma{\hat{m}^{(i)}}/{(\sqrt{\hat{v}^{(i)}} + \epsilon)},
\end{align}
where $\gamma$ is a learning rate, and $\beta_1$, $\beta_2$, and $\epsilon$ are standard parameters of the Adam optimizer (see \citealpAPP{kingma2014adamAPP} for further details).
Using this modified version of Adam, which we call ``decoupled Adam'', we considered two variants of KFAC: (i) ``decoupled KFAC'', where we fix $\lambda_m =1$ and vary $\lambda_v$~(\Cref{fig:by_lambda}C), and (ii) ``reverse decoupled'', where we  fix $\lambda_v=1$ and vary $\lambda_m$~(\Cref{fig:by_lambda}D).
We found that ``decoupled KFAC'' performed well for large $\lambda_v$, suggesting that it is sufficient to overcount the prior in the Adam preconditioner without changing the gradient estimate (\Cref{fig:by_lambda}C).
``Reverse decoupled'' also partly overcame the catastrophic forgetting for high $\lambda_m$, but performance was worse than for either NCL, vanilla Adam, or decoupled Adam (\Cref{fig:by_lambda}D).
These results support our hypothesis that the increased performance of KFAC for high $\lambda$ is due in part to the changes in the gradient preconditioner.
To further highlight how the preconditioning in Adam relates to the trust region optimization employed by NCL, we computed the scaled KL divergence between the Adam preconditioner and the diagonal of the Kronecker-factored prior precision matrix $\b{\Lambda}_{k-1}$ at the end of training on task $k$.
We found that the Adam preconditioner increasingly resembled $\b{\Lambda}_{k-1}$, the preconditioner used by NCL, as $\lambda$ increased (\Cref{fig:kfac_lambda_prec}).

In summary, our results suggest that preconditioning with $\b{\Lambda}_{k-1}$ in NCL may mitigate the need to overcount the prior when using weight regularization for continual learning.
Additionally, such preconditioning to encourage parameter updates that retain good performance on previous tasks also appears to be a major contributing factor to the success of weight regularization with a high value of $\lambda$ when using Adam for optimization.

\subsection{Hyperparameter optimizations}
\label{sec:hp_opt}

\paragraph{Feedforward networks}
For the experiments with feedforward networks, we performed hyperparameter optimizations by searching over the following parameter ranges (all on a log-scale): $c$ in SI from $10^{-5}$ to $10^8$, $\lambda$ in EWC and KFAC from $10^{-4}$ to $10^{14}$, $\alpha$ in OWM from $10^{-12}$ to $10^6$, and $p_{w}^{-2}$ in NCL from $10^2$ to $10^{11}$. 
The hyperparameter grid searches were performed using a random seed not included during the evaluation. 
The selected hyperparameter values for each experiment are reported in \Cref{tab:hp}.

\begin{table}[h]
  \parbox{0.79\textwidth}{
    \begin{center}
      \begin{small}
        \begin{tabular}{lp{1.1cm}p{1.1cm}p{1.1cm}p{1.1cm}p{1.1cm}p{1.1cm}} \toprule
               & \multicolumn{3}{c}{\textbf{Split MNIST}} & \multicolumn{3}{c}{\textbf{Split CIFAR-100}}                                                                     \\
               & \textbf{Task}                            & \textbf{Domain}                              & \textbf{Class} & \textbf{Task} & \textbf{Domain} & \textbf{Class} \\ \midrule \midrule
          SI ($c$)   & $10^{0}$                                 & $10^{5}$                                     & $10^{7}$       & $10^{2}$      & $10^{3}$        & $10^{6}$       \\
          EWC ($\lambda$) & $10^{8}$                                 & $10^{9}$                                     & $10^{13}$      & $10^{7}$      & $10^{5}$        & $10^{-3}$      \\
          KFAC ($\lambda$) & $10^{10}$                                & $10^{5}$                                     & $10^{4}$       & $10^{5}$      & $10^{3}$        & $10^{10}$      \\
          OWM ($\alpha$) & $10^{-2}$                                & $10^{-5}$                                    & $10^{-4}$      & $10^{-2}$     & $10^{-2}$       & $10^{-4}$      \\
          NCL ($p_w^{-2}$) & $10^{3}$                                 & $10^{7}$                                     & $10^{9}$       & $10^{3}$      & $10^{7}$        & $10^{8}$       \\
          \bottomrule
        \end{tabular}
      \end{small}
    \end{center}
  }
  \parbox{0.20\textwidth}{
    \caption{\label{tab:hp}Selected hyperparameter values for all compared methods on the experiments with feedforward networks.}
  }
\end{table}

\paragraph{RNNs}
For the experiments with RNNs, we optimized over the parameter $\alpha$ used to invert the approximate Fisher matrices in the projection-based methods (NCL, OWM and DOWM) or over the parameter $\lambda$ used to scale the importance of the prior for weight regularization (KFAC).

\begin{figure}[!t]
    \centering
    \includegraphics[width = 0.79\textwidth, trim={0 0 0 0}, clip=true]{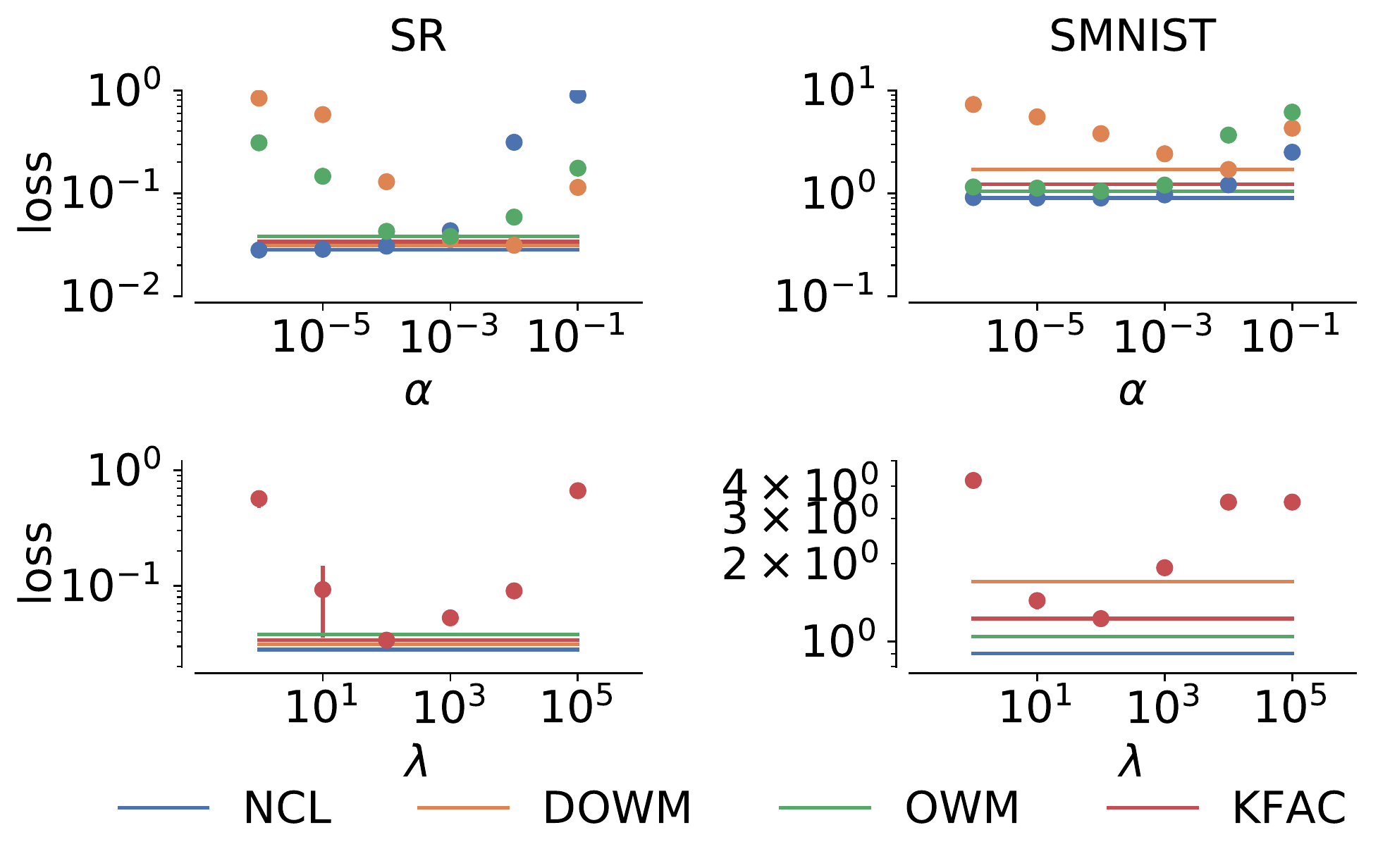}
    \caption{\label{fig:hp_opt}
    {\bfseries Hyperparameter optimization for RNNs.}
    {\bfseries (A)}~Comparison of the average loss across tasks on the stimulus-response task set as a function of $\alpha$ for the projection-based methods (top panel) and as a function of $\lambda$ for KFAC (bottom panel).
    Circles and error bars indicate mean and s.e.m. across 5 random seeds.
    Horizontal lines indicate the optimal value for each method.
    {\bfseries (B)}~As in (A), now for the SMNIST task set.
    }
    \vspace*{-1em}
\end{figure}

For KFAC, we found that the performance was very sensitive to the value of $\lambda$ across all tasks sets, and in particular that $\lambda = 1$ performed poorly.
In the projection-based methods, $\alpha$ can be seen as evening out the learnings rates between directions that are otherwise constrained by the projection matrices, and indeed standard gradient descent is recovered as $\alpha \rightarrow \infty$ (on the Laplace objective for NCL and on $\ell_k$ for OWM/DOWM).
We found that NCL in general outperformed the other projection-based methods with less sensitivity to the regularization parameter $\alpha$.
DOWM was particularly sensitive to $\alpha$ and required a relatively high value of this parameter to balance its otherwise conservative projection matrices (\Cref{sec:proj}).
Here it is also worth noting that there is an extensive literature on how a parameter equivalent to $\alpha$ can be dynamically adjusted when doing standard natural gradient descent using the Fisher matrix for the current loss (see \citealpAPP{martens2014newAPP} for an overview).
While this has not been explored in the context of projection-based continual learning, it could be interesting to combine these projection based methods with Tikhonov dampening \citepAPP{tikhonov1943stabilityAPP} in future work to automatically adjust $\alpha$.

We generally report results in the main text and appendix using the optimal hyperparameter settings for each method unless otherwise noted.
However, $\alpha = 10^{-5}$ was used for both NCL and Laplace-DOWM in \CrefNoLink{fig:low_cap}C to compare the qualitative behavior of the two different Fisher approximations without the confound of a large learning rate in directions otherwise deemed ``important'' by the approximation.

\subsection{Numerical results of experiments with feedforward networks}
\label{sec:numerical_results}
To facilitate comparison to our results, here we provide a table with the numerical results (\Cref{tab:numerical_results}) of the experiments with feedforward networks reported in \Cref{fig:ff} of the main text.

\begin{table}[h!]
  \begin{center}
  \begin{small}
  \caption{\label{tab:numerical_results}Numerical results for the experiments with feedforward networks, corresponding to \Cref{fig:ff} in the main text. Reported is the test accuracy (as \%, averaged over all tasks) after training on all tasks. Each experiment was performed either 20 (split MNIST) or 10 (split CIFAR-100) times with different random seeds, and we report the mean ($\pm$ standard error) across seeds.}
  \vskip 0.1in
  \begin{tabular}{lp{1.5cm}p{1.5cm}p{1.5cm}p{0.0001cm}p{1.5cm}p{1.5cm}p{1.5cm}} \toprule
    & \multicolumn{3}{c}{\textbf{Split MNIST}} && \multicolumn{3}{c}{\textbf{Split CIFAR-100}} \\                                                     \textbf{Method} & \textbf{~~~~Task} & \textbf{~~Domain} & \textbf{~~~~Class} && \textbf{~~~~Task} & \textbf{~~Domain} & \textbf{~~~~Class} \\ \midrule \midrule
    \it None & \it 81.58 \scriptsize{$\pm$1.64} & \it 59.47 \scriptsize{$\pm$1.71} & \it 19.88 \scriptsize{$\pm$0.02}  && \it 61.43 \scriptsize{$\pm$0.36}  & \it 18.42 \scriptsize{$\pm$0.33}  & \it ~~7.71 \scriptsize{$\pm$0.18} \\
    \it Joint & \it 99.69 \scriptsize{$\pm$0.02} & \it 98.69 \scriptsize{$\pm$0.04} & \it 98.32 \scriptsize{$\pm$0.05}  && \it 78.78 \scriptsize{$\pm$0.25}  & \it 46.85 \scriptsize{$\pm$0.51}  & \it 49.78 \scriptsize{$\pm$0.21} \\ \midrule
    SI & 97.24 \scriptsize{$\pm$0.55}  & 65.20 \scriptsize{$\pm$1.48} & 21.40 \scriptsize{$\pm$1.30} && 74.84 \scriptsize{$\pm$0.39}  & 22.58 \scriptsize{$\pm$0.37} & ~~7.02 \scriptsize{$\pm$1.04} \\
    EWC & 98.67 \scriptsize{$\pm$0.22}  & 63.44 \scriptsize{$\pm$1.70} & 20.08 \scriptsize{$\pm$0.16} && 75.38 \scriptsize{$\pm$0.24}  & 19.97 \scriptsize{$\pm$0.44} & ~~7.74 \scriptsize{$\pm$0.18} \\
    KFAC & 99.04 \scriptsize{$\pm$0.10}  & 67.86 \scriptsize{$\pm$1.33} & 19.99 \scriptsize{$\pm$0.04} && 76.61 \scriptsize{$\pm$0.23}  & 26.57 \scriptsize{$\pm$0.66} & ~~7.59 \scriptsize{$\pm$0.17} \\
    OWM & 99.36 \scriptsize{$\pm$0.05}  & 87.46 \scriptsize{$\pm$0.74} & 80.73 \scriptsize{$\pm$1.11} && 77.07 \scriptsize{$\pm$0.27}  & 28.51 \scriptsize{$\pm$0.30} & 29.23 \scriptsize{$\pm$0.51} \\
    NCL (no opt) & 99.53 \scriptsize{$\pm$0.03}  & 84.9 \scriptsize{$\pm$1.06} & 47.49 \scriptsize{$\pm$0.84} && 77.88 \scriptsize{$\pm$0.26}  & 32.81 \scriptsize{$\pm$0.38} & 16.63 \scriptsize{$\pm$0.34} \\
    NCL & 99.55 \scriptsize{$\pm$0.03}  & 91.48 \scriptsize{$\pm$0.64} & 69.31 \scriptsize{$\pm$1.65} && 78.38 \scriptsize{$\pm$0.27}  & 38.79 \scriptsize{$\pm$0.24} & 26.36 \scriptsize{$\pm$1.09} \\ \bottomrule
  \end{tabular}
  \end{small}
  \end{center}
\end{table}

\newpage 

\subsection{SMNIST dynamics with DOWM and replay}
\label{sec:dynamics_dowm}

\begin{figure}[!t]
    \centering
    \includegraphics[width = 0.98\textwidth, trim={0 0 0 0}, clip=true]{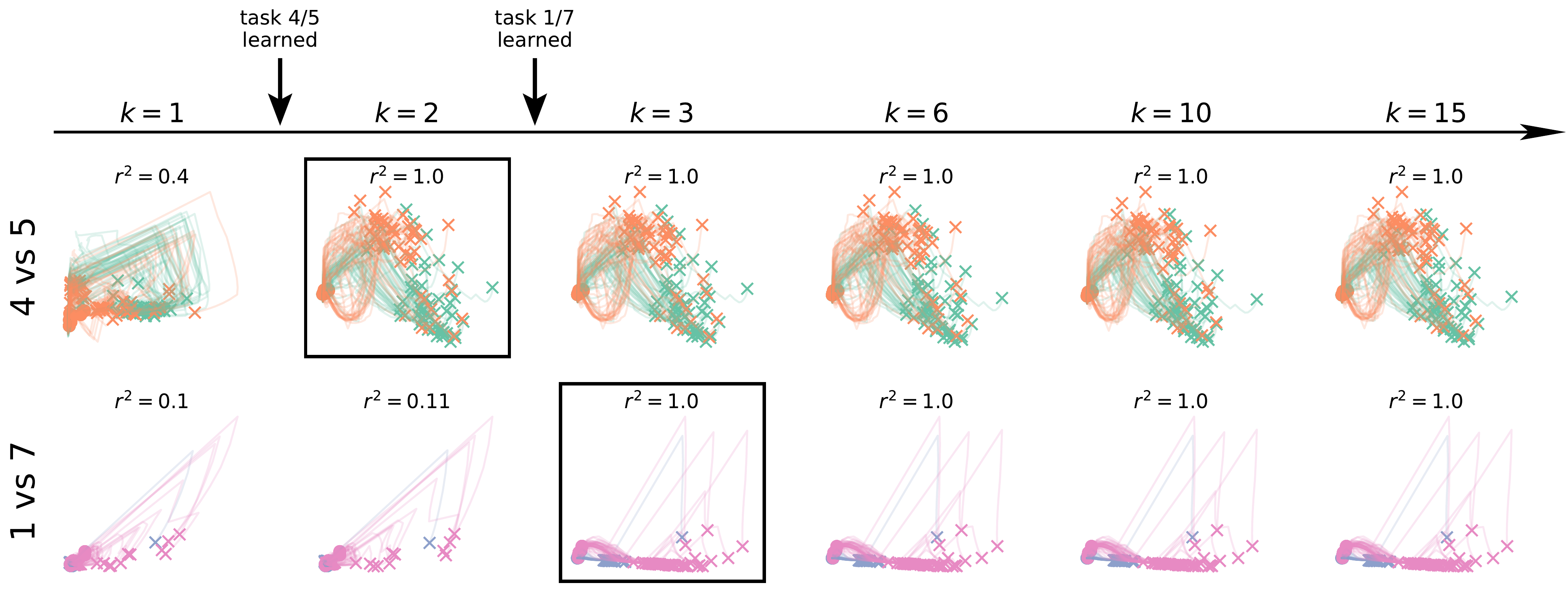}
    \caption{\label{fig:smnist_dyn_dowm}
        {\bfseries Latent dynamics during SMNIST.}
        We considered two example tasks, 4 vs 5 (top) and 1 vs 7 (bottom).
        For each task, we simulated the response of a network trained by DOWM to 100 digits drawn from that task distribution at different times during learning.
        We then fitted a factor analysis model for each example task to the response of the network right after the correponding task had been learned (squares; $k = 2$ and $k=3$ respectively).
        We used this model to project the responses at different times during learning into a common latent space for each example task.
        For both example tasks, the network initially exhibited variable dynamics with no clear separation of inputs and subsequently acquired stable dynamics after learning to solve the task.
        The $r^2$ values above each plot indicate the similarity of neural population activity with that collected immediately after learning the corresponding task, quantified across all neurons (not just the 2D projection).
        }
    \vspace*{-1em}
\end{figure}

In this section, we investigate the latent dynamics of a network trained by DOWM with $\alpha = 0.001$ (c.f. the analysis in \CrefNoLink{subsec:dynamics} for NCL).
Here we found that the task-associated recurrent dynamics for a given task were more stable after learning the corresponding task than in networks trained with NCL.
Indeed, the DOWM networks exhibited near-zero drift for early tasks even after learning all 15 tasks (\Cref{fig:smnist_dyn_dowm}).
However, DOWM also learned representations that were less well-separated after the first 1-2 classification tasks (\Cref{fig:smnist_dyn_dowm}, bottom) than those learned by NCL.
This is consistent with our results in \CrefNoLink{subsec:smnist} where DOWM exhibited high performance on the first task even after learning all 15 tasks, but performed less well on later tasks (\Cref{fig:smnist_dyn_dowm}).
These results may be explained by the observation that DOWM tends to overestimate the number of dimensions that are important for learned tasks (\CrefNoLink{subsec:smnist}) and thus projects out too many dimensions in the parameter updates when learning new tasks.

In the context of biological networks, it is unlikely that the brain remembers previous tasks in a way that causes it to lose the capacity to learn new tasks.
However, it is also not clear how the balance between capacity and task complexity plays out in the mammalian brain, which on the one hand has many orders of magnitude more neurons than the networks analyzed here, but on the other hand also learns more behaviors that are more complex than the problems studied in this work.
In networks where capacity is not a concern, it may in fact be desirable to employ a strategy similar to that of DOWM --- projecting out more dimensions in the parameter updates than is strictly necessary --- so as to avoid forgetting in the face of the inevitable noise and turnover of e.g. synapses and cells in biological systems.

To compare with NCL and DOWM which ensure continual learning by regularizing the parameters of the network or the changes in these parameters, we also considered a network that used replay of past data to avoid forgetting.
We trained the network using a simple implementation of replay where the learner estimates the task specific loss $\ell_k(\theta)$ as in \CrefNoLink{sec:method}.
In addition, the network gets to `replay' a set of examples $\{ \b{x}^{(k')}, \b{y}^{(k')} \}_{k' < k}$ from previous tasks at every iteration to estimate the expected loss on earlier tasks
\begin{equation}
  \label{eq:replay}
  \ell_{<k}(\b{\theta}) = \frac{1}{k-1} \sum_{k' = 1}^{k-1} \mathbb{E} \left [ \sum_t \log p(\b{y}^{(k')}_t | \b{C} \b{r}^{(k')}_t) \right ].
\end{equation}
The parameters are then updated as
\begin{equation}
  \b{\theta} \leftarrow \b{\theta} - \gamma \left [ \frac{1}{k} \nabla_\theta \ell_k(\b{\theta}) + \frac{k-1}{k} \nabla_\theta \ell_{<k}(\b{\theta}) \right].
\end{equation}
Note that while we explicitly replayed examples drawn from the true data distribution for previous tasks, these examples could instead be drawn from a generative model that is learned in a continual fashion together with the discriminative model \citepAPP{van2018generativeAPP, van2020brainAPP}.

\begin{figure}[!t]
    \centering
    \includegraphics[width = 0.98\textwidth, trim={0 0 0 0}, clip=true]{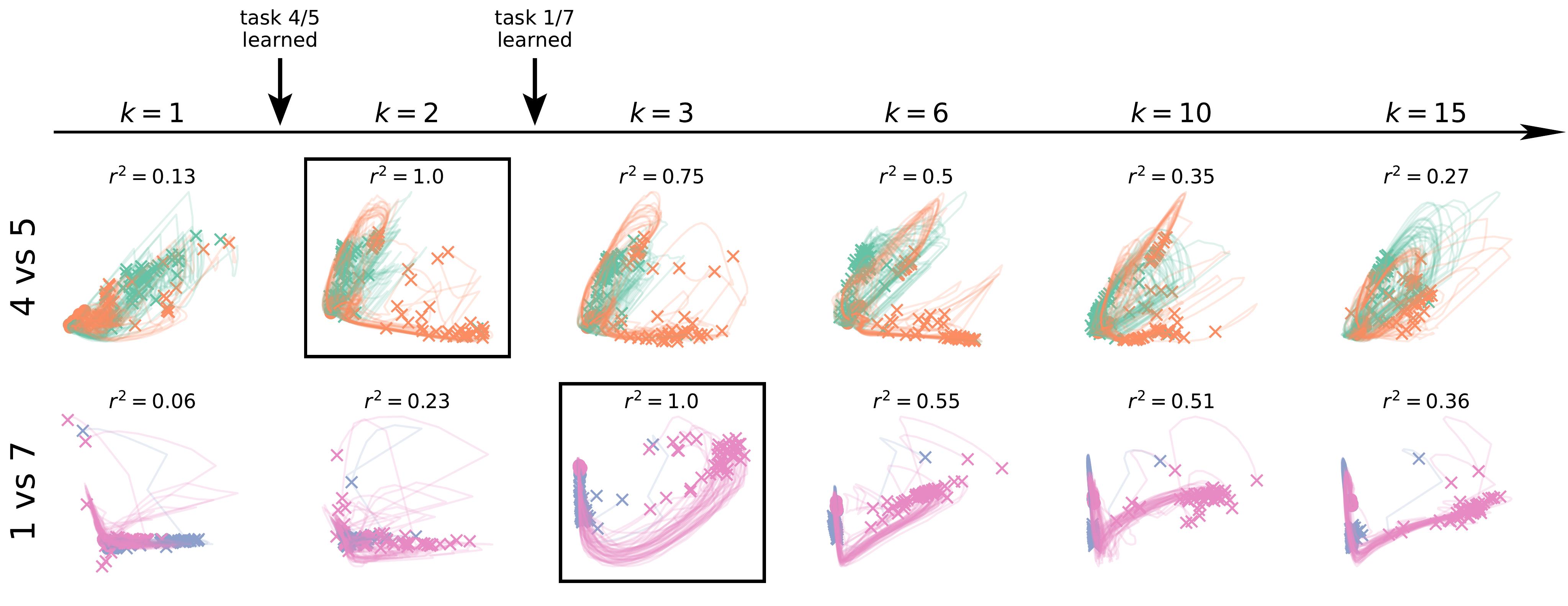}
    \caption{\label{fig:smnist_dyn_replay}
        {\bfseries Latent dynamics during SMNIST with replay.}
        We considered two example tasks, 4 vs 5 (top) and 1 vs 7 (bottom).
        For each task, we simulated the response of a network trained with replay (see main text for details) to 100 digits drawn from that task distribution at different times during learning.
        We then proceeded to analyze the latent space and representational stability as in \Cref{fig:smnist_dyn_dowm}.
        }
    \vspace*{-1em}
\end{figure}

In contrast to the stable task representations learned by DOWM and NCL, continual learning with replay led to task-representations that exhibited a continuous drift after the initial task acquisition (\Cref{fig:smnist_dyn_replay}).
We can understand this by noting that parameter-based continual learning assigns a privileged status to the parameter set $\b{\mu}_k$ used when the task is first learned, while methods using replay, generative replay, or other functional regularization methods are invariant to parameter changes that do not affect the functional mapping $f_\theta(x)$.
This is interesting since a range of studies in the neuroscience literature have investigated the stability of neural representations with some indicating stable representations \citepAPP{chestek2007singleAPP,dhawale2017automatedAPP,katlowitz2018stableAPP,jensen2021longAPP} and others drifting representations \citepAPP{liberti2016unstableAPP, driscoll2017dynamicAPP, schoonover2021representationalAPP, rokni2007motorAPP}.
It is thus possible that these differences in experimental findings could reflect differences between animals, tasks and neural circuits in how biological continual learning is implemented.
In particular, stable representations could result from selective stabilitization of individual synapses as in parameter regularization methods for continual learning, and drifting representations could result from regularizing a functional mapping using generative replay.

\section{NCL for variational continual learning}
\label{sec:vcl}
\paragraph{Online variational inference}

In variational continual learning \citepAPP{nguyen2017variationalAPP}, the posterior $p(\b{\theta} | \mathcal{D}_k, \b{\phi}_{k-1})$ is approximated with a Gaussian variational distribution $q(\b{\theta}_k; \b{\phi}_k) = \mathcal{N}(\b{\theta}_k; \b{\mu}, \b{\Sigma}_k)$, where $\b{\phi}_k = (\b{\mu}_k, \b{\Sigma}_k)$.
We then treat $\b{\mu}_k$ and $\b{\Sigma}_k$ as variational parameters and minimize the KL-divergence between $q(\b{\theta}_k; \b{\phi}_k)$ and the approximate posterior $p(\b{\theta}|\mathcal{D}_k, \b{\phi}_{k}) \propto q(\b{\theta}; \b{\mu}_{k-1}, \b{\Sigma}_{k-1}) p(\mathcal{D}_k|\b{\theta})$:
\begin{align}
	\text{KL}                                     
	\left (                                       
	q(\b{\theta};\b{\mu}_k, \b{\Sigma}_k) ||      
	\frac{1}{Z_k}                                 
	q(\b{\theta};\b{\mu}_{k-1}, \b{\Sigma}_{k-1}) 
	p(\mathcal{D}_k|\b{\theta})                   
	\right )                                      
\end{align}
with respect to $\b{\mu}_k$ and $\b{\Sigma}_k$.
This is equivalent to maximizing the evidence lower-bound (ELBO):
\begin{equation}
	\mathcal{L}(\b{\mu}_k, \b{\Sigma}_k)=
	\mathbb{E}_{\b{q}(\b{\theta};\b{\mu}_k, \b{\Sigma}_k)}
	\left [
		\log p(\mathcal{D}_k|\b{\theta})
	\right ]
	-
	\text{KL}
	\left (
	q(\b{\theta};\b{\mu}_k, \b{\Sigma}_k) ||
	q(\b{\theta};\b{\mu}_{k-1}, \b{\Sigma}_{k-1})
	\right ),
	\label{eq:elbo}
\end{equation}
to the data log likelihood
\begin{align}
	\log p(\mathcal{D}_k|\b{\phi}_{k-1})      
	=                                         
	\log \int p(\mathcal{D}_k | \b{\theta})   
	q(\b{\theta}; \b{\phi}_{k-1}) d \b{\theta} 
	\geq                                      
	\mathcal{L}                               
\end{align}
with $q(\b{\theta};\b{\phi}_{k-1})$ as the `prior' for task $k$.

Maximizing $\mathcal{L}$ requires the computation of both the first likelihood term and the second KL term in \Cref{eq:elbo}.
While the second term can be computed analytically, the first term is intractable for general likelihoods $p(\mathcal{D}_k|\b{\theta})$.
To address this, \citetAPP{nguyen2017variationalAPP} estimate this likelihood term using Monte Carlo sampling:
\begin{equation}
	\mathbb{E}_{q(\b{\theta};\b{\phi}_k)} \left [ \log p(\mathcal{D}_k | \b{\theta}) \right  ]
	\approx
	\frac{1}{K} \sum_i
	\log p(\mathcal{D}_k | \b{\theta}_i),
\end{equation}
where $\{\b{\theta}_i\}_{i=1}^M \sim q(\b{\theta};\b{\phi}_k)$ are drawn from the variational posterior via the reparameterization trick.
This allows direct optimization of the variational parameters $\b{\mu}_k$ and $\b{\Sigma}_k$. 
To make the method scale to large models with potentially millions of parameters, \citetAPP{nguyen2017variationalAPP} also make a mean-field approximation to the posterior
\begin{equation}
	q(\b{\theta}; \b{\phi}_k) =
	\mathcal{N}(\b{\theta}; \b{\mu}_k, \text{diag}(\b{\sigma}_k)).
\end{equation}
\paragraph{Natural variational continual learning} 
We now propose an alternative approach to maximizing $\mathcal{L}$ with respect to $\b{\phi}_k = (\b{\mu}_k, \b{\Lambda}_k)$ within the NCL framework, where $\b{\Lambda}_k = \b{\Sigma}_k^{-1}$ is the precision matrix of $q$ at step $k$. 
We again solve a trust-region subproblem to find the optimal parameter updates for $\b{\mu}_k$ and $\b{\Lambda}_k$:
\begin{align}
	  &                  
	\b{\Delta}_{\b{\mu}_k},
	\b{\Delta}_{\b{\Lambda}_k}
	=
	\argmin_{
	\b{\Delta}_{\b{\mu}_k},
	\b{\Delta}_{\b{\Lambda}_k}
	}
	\mathcal{L}(\b{\mu}_k, \b{\Lambda}_k) 
	+ 
	\nabla_{\b{\mu}_k}\mathcal{L}^\top \b{\Delta}_{\b{\mu}_k}
	+ 
	\nabla_{\b{\Lambda}_k}\mathcal{L}^\top \b{\Delta}_{\b{\Lambda}_k}\\
	  & \text{such that} 
	\quad
	\mathcal{C}(\b{\Delta}_{\b{\mu}_k}, \b{\Delta}_{\b{\Lambda}_k})
	\leq r^2,
\end{align}
where 
\begin{equation}
	\mathcal{C}(\b{\Delta}_{\b{\mu}_k}, \b{\Delta}_{\b{\Lambda}_k})
	=
	\frac{1}{2}\b{\Delta}_{\b{\mu}_k}^\top \b{\Lambda}_{k-1} \b{\Delta}_{\b{\mu}_k}
	+
	\frac{1}{4}\vect(\b{\Delta}_{\b{\Lambda}_k})^\top
	(
	\b{\Lambda}_{k}^{-1}
	\otimes
	\b{\Lambda}_{k}^{-1}
	)
	\vect(\b{\Delta}_{\b{\Lambda}_k}).
\end{equation}
The solution to this optimization problem is given by:
\begin{align}
	\b{\Delta}_{\b{\mu}_k} 
	  & = \b{\Lambda}_{k-1}^{-1} \nabla_{\b{\mu}_k}\mathcal{L} \\
	\b{\Delta}_{\b{\Lambda}_k} 
	  & =                                                      
	2 \b{\Lambda}_{k} 
	\nabla_{\b{\Lambda}_k}\mathcal{L}
	\b{\Lambda}_{k}.
\end{align}
To compute $\nabla_{\b{\mu}_k} \mathcal{L}$ and $\nabla_{\b{\Lambda_k}} \mathcal{L}$, we make use of  the following identities~\citepAPP{opper2009variationalAPP}:
\begin{align}
	\nabla_{\b{\mu}}~
	\mathbb{E}_{\b{\theta} \sim \mathcal{N}(\b{\theta}; \b{\mu}, \b{\Sigma})}
	\left [
	f(\b{\theta})
	\right ]
	  & = 
	\mathbb{E}_{\b{\theta} \sim \mathcal{N}(\b{\theta}; \b{\mu}, \b{\Sigma})}
	\left [
	\nabla_{\b{\theta}} f(\b{\theta})
	\right ]
	\label{eq:nabla_mu} \\
	\nabla_{\b{\Sigma}}~
	\mathbb{E}_{\b{\theta} \sim \mathcal{N}(\b{\theta}; \b{\mu}, \b{\Sigma})}
	\left [
	f(\b{\theta})
	\right ]
	  & = 
	\frac{1}{2} \mathbb{E}_{\b{\theta} \sim \mathcal{N}(\b{\theta}; \b{\mu}, \b{\Sigma})}
	\left [
	\nabla^2_{\b{\theta}} f(\b{\theta})
	\right ].
	\label{eq:nabla_sigma}
\end{align}
Applying these identities to compute the gradients of $\mathcal{L}$ (\Cref{eq:elbo}), we find
\begin{align}
	\nabla_{\b{\mu}_k} \mathcal{L}    & = 
	\mathbb{E}_{\b{\theta} \sim q(\b{\theta};\b{\phi}_k)}
	\left [
	\nabla_{\b{\theta}} \log p(\mathcal{D}_k|\b{\theta})
	- \b{\Lambda}_{k-1}(\b{\theta}_k - \b{\mu}_{k-1})
	\right ]                          \\
	\nabla_{\b{\Sigma}_k} \mathcal{L} & = 
	\frac{1}{2} \mathbb{E}_{\b{\theta} \sim q(\b{\theta};\b{\phi}_k)}
	\left [
	\nabla_{\b{\theta}}^2 \log p(\mathcal{D}_k|\b{\theta})
	- \b{\Lambda}_{k-1}
	+ \b{\Lambda}_k
	\right ]
	.
\end{align}
Using the fact that $d\b{\Lambda}_k =  - \b{\Lambda}^{-1}_kd\b{\Sigma}_k \b{\Lambda}^{-1}_k$, we have
\begin{align}
	\nabla_{\b{\Lambda}_k} \mathcal{L}
	  & = - \b{\Lambda}_k^{-1} \nabla_{\b{\Sigma}_k} \mathcal{L} \b{\Lambda}_k^{-1} \\
	  & =                                                                           
	-\frac{1}{2}
	\b{\Lambda}_k^{-1}
	\mathbb{E}_{\b{\theta} \sim q(\b{\theta};\b{\phi}_k)}
	\left [
	\nabla_{\b{\theta}}^2 \log p(\mathcal{D}_k|\b{\theta})
	- \b{\Lambda}_{k-1}
	+ \b{\Lambda}_k
	\right ]
	\b{\Lambda}_k^{-1}.
\end{align}

This suggests that we can compute $\b{\Delta}_{\b{\mu}_k}$ and $\b{\Delta}_{\b{\Lambda}_k}$ as:
\begin{align}
	\b{\Delta}_{\b{\mu}_k} 
	  & = \b{\Lambda}_{k-1}^{-1} 
	\mathbb{E}_{\b{\theta} \sim q(\b{\theta};\b{\phi}_k)}
	\left [ 
	\nabla_{\b{\theta}} \log p(\mathcal{D}_k | \b{\theta}_k)
	\right ]
	- (\b{\mu}_k - \b{\mu}_{k-1})
	\\
	\b{\Delta}_{\b{\Lambda}_k} 
	  & =                        
	\b{\Lambda}_{k-1}
	-
	\b{\Lambda}_{k}
	-
	\mathbb{E}_{\b{\theta} \sim q(\b{\theta};\b{\phi}_k)}
	\left [
	\nabla_{\b{\theta}}^2 \log p(\mathcal{D}_k|\b{\theta})
	\right ].
\end{align}
This gives the following update rule at learning iteration $i$ during task $k$:
\begin{align}
	\b{\mu}_k^{(i+1)}
	  & = 
	(1- \beta)
	\b{\mu}_k^{(i)}
	+
	\beta
	\left [
	\b{\mu}_{k-1} + \b{\Lambda}_{k-1}^{-1}
	\mathbb{E}_{\b{\theta} \sim q(\b{\theta};\b{\phi}_k^{(i)})}
	\left [
	\nabla_{\b{\theta}} \log p(\mathcal{D}_k | \b{\theta})
	\right ]
	\right ]
	\label{eq:update_rule_mu}
	\\
	\b{\Lambda}_k^{(i+1)}
	  & = 
	(1- \beta)
	\b{\Lambda}_{k}^{(i)}
	+
	\beta
	\left [
	\b{\Lambda}_{k-1}
	-
	\mathbb{E}_{\b{\theta} \sim q(\b{\theta};\b{\phi}_k^{(i)})}
	\left [
	\nabla^2_{\b{\theta}} \log p(\mathcal{D}_k | \b{\theta})
	\right ]
	\right ]
	\label{eq:update_rule_lambda}
	,
\end{align}
Note that this update rule is equivalent to preconditioning the gradients $\nabla_{\b{\mu}_k} \mathcal{L}$ and $\nabla_{\b{\Lambda}_k} \mathcal{L}$ with $\b{\Lambda}_{k-1}^{-1}$ and
$\b{\Lambda}_{k} \otimes \b{\Lambda}_{k}$ respectively.

As for the online Laplace approximation (\CrefNoLink{subsec:bcl}), one of the main difficulties of implementing the update rule described in \Cref{eq:update_rule_mu} and \Cref{eq:update_rule_lambda} is that it is impractical to compute and store the Hessian of the negative log-likelihood for large models. 
Furthermore, we need $\b{\Lambda}_k^{-1}$ to remain PSD which is not guaranteed as the Hessian is not necessarily PSD.
In practice we therefore again approximate the Hessian with the Fisher-information matrix:
\begin{equation}
	H_k = -\mathbb{E}
	\left [
		\nabla_{\theta}^2 \log p(\b{\theta})
	\right ]
	\approx
	F_k =
	\mathbb{E}_{\hat{\mathcal{D}}_k \sim p(\mathcal{D}_k|\b{\theta})}
	\left [
		\nabla_{\theta} \log p(\hat{\mathcal{D}}_k|\b{\theta})
		\nabla_{\theta} \log p(\hat{\mathcal{D}}_k|\b{\theta})^{\top}
	\right ].
\end{equation}
As in \CrefNoLink{subsec:ngcl} we use a Kronecker factored approximation to the FIM for computational tractability.
With these approximations, we arrive at the learning rule:
\begin{align}
	\b{\mu}_k^{(i+1)}
	  & = 
	(1- \beta)
	\b{\mu}_k^{(i)}
	+
	\beta
	\left [
	\b{\mu}_{k-1} + \b{\Lambda}_{k-1}^{-1}
	\mathbb{E}_{\b{\theta} \sim q(\b{\theta};\b{\phi}_k^{(i)})}
	\left [
	\nabla_{\b{\theta}} \log p(\mathcal{D}_k | \b{\theta})
	\right ]
	\right ],
	\label{eq:ngcl_mu}
	\\
	\b{\Lambda}_k^{(i+1)}
	  & = 
	(1- \beta)
	\b{\Lambda}_{k}^{(i)}
	+
	\beta
	\left [
	\b{\Lambda}_{k-1}
	+
	\mathbb{E}_{\b{\theta} \sim q(\b{\theta};\b{\phi}_k^{(i)})}
	\left [
	F_k(\b{\theta})
	\right ]
	\right ].
	\label{eq:ngcl_lambd}
\end{align}

These update rules define the `natural variational continual learning' (NVCL) algorithm which is the variational equivalent of the Laplace algorithm derived in \CrefNoLink{subsec:ngcl} and used in \CrefNoLink{sec:exps}.

\paragraph{Experiments}

\begin{figure}[!t]
    \parbox{0.55\textwidth}{
        \centering
        \includegraphics[width = 0.53\textwidth]{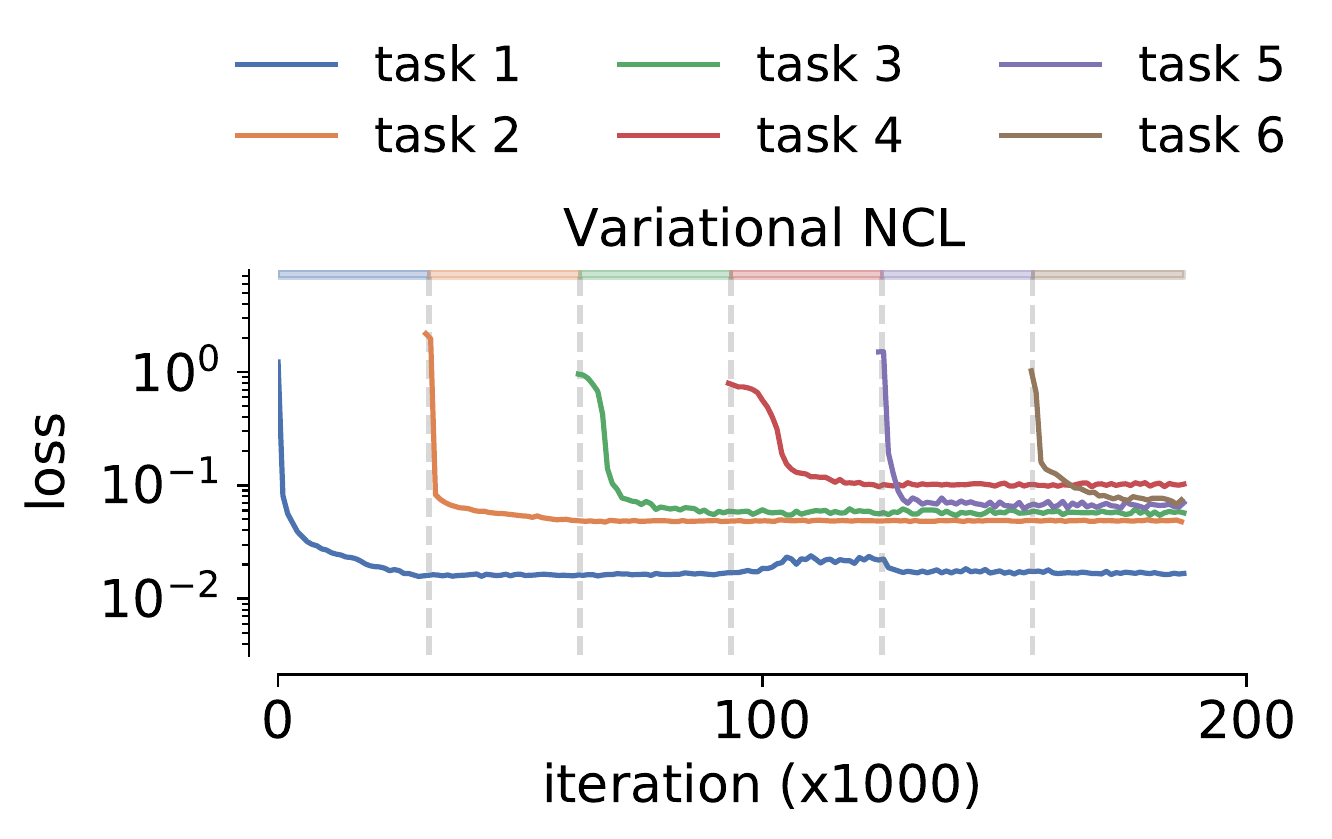}
    }
    \parbox{0.43\textwidth}{
        \caption{\label{fig:ryang_variational_ncl}
            {\bfseries Natural VCL applied to the stimulus-response task.}
            Evolution of the loss during training for each of the six stimulus-response tasks using NVCL.
        }}
\end{figure}

To understand how \Crefrange{eq:ngcl_mu}{eq:ngcl_lambd} encourage continual learning, we note that the first two terms of \Cref{eq:ngcl_mu} urge the new parameters $\b{\mu}_k$ to stay close to $\b{\mu}_{k-1}$.
The third term of \Cref{eq:ngcl_mu} improves the average performance of the learner on task $k$ by moving $\b{\mu}_k$ along $\b{\Lambda}_{k-1}^{-1} p(\mathcal{D}_k|\b{\theta})$.
This is a valid search direction because $\b{\Lambda}_{k-1}^{-1} = \b{\Sigma}_k$ is the covariance of $q(\b{\theta};\b{\phi}_{k-1})$ and is thus positive semi-definite (PSD).
Importantly, the preconditioner $\b{\Lambda}_{k-1}^{-1}$ ensures that $\b{\mu}_k$ changes primarily along ``flat'' directions of $q(\b{\theta};\b{\phi}_{k-1})$.
This in turn encourages $q(\b{\theta};\b{\phi}_k)$ to stay close to $q(\b{\theta};\b{\phi}_{k-1})$ in the KL sense.
In \Cref{eq:ngcl_lambd}, the first two terms again encourage $\b{\Lambda}_k$ to remain close to $\b{\Lambda}_{k-1}$.
The third term in \Cref{eq:ngcl_lambd} updates the precision matrix of the approximate posterior with the average Fisher matrix for task $k$.
This encourages the curvature of the approximate posterior to be similar to that of the loss landscape of task $k$, and thus (at least locally) parameters that have similar performance on the task will have similar probabilities under the approximate posterior.

To test the natural VCL algorithm, we applied it to the stimulus-response task set considered in \CrefNoLink{subsec:ryang} using an RNN with 256 units.
Similar to the Laplace version of NCL, we found that NVCL was capable of solving all six tasks without forgetting (\Cref{fig:ryang_variational_ncl}).
While this can be seen as a proof-of-principle that our natural VCL algorithm works, we leave more extensive comparisons between the variational and Laplace algorithms for future work.

\paragraph{Related work} 
Previous studies have proposed the use of variants of natural gradient descent to optimize the variational continual learning objective~\citepAPP{tseran2018naturalAPP,osawa2019practicalAPP}.
The key differences between the method proposed in this section and previous methods are two-fold: (i) we precondition the gradient updates on task $k$ with $\b{\Lambda}_{k-1}^{-1}$ as opposed to $\b{\Lambda}_k^{-1}$ as is done in prior work, and (ii) we estimate the Fisher matrix on each task by drawing samples from the model distribution as opposed to the empirical distribution as is the case in~\citetAPP{tseran2018naturalAPP,osawa2019practicalAPP}.
It has previously been argued that drawing from the model distribution instead of using the `empirical' Fisher matrix is important to retain the desirable properties of natural gradient descent \citep{kunstner2019limitationsAPP}.

\section{Details of toy example in schematic}
\label{sec:toy}
\begin{figure}[t]
    \centering
    \includegraphics[width = 0.7\textwidth, trim={0 0 0 0}, clip=true]{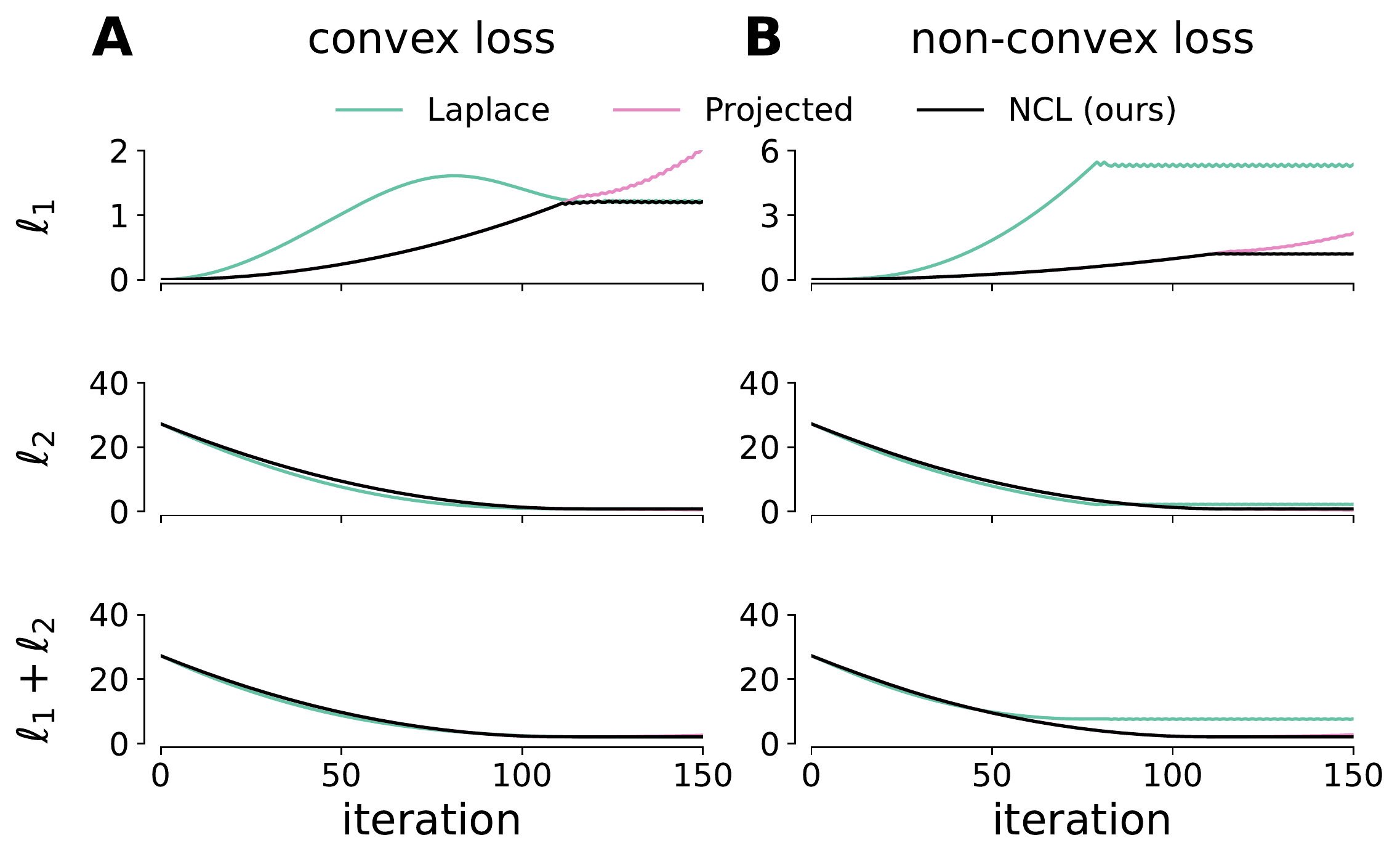}
    \caption{\label{fig:schem_losses}
        {\bfseries Losses on toy optimization problem.}
        {\bfseries (A)}~Loss as a function of optimization step on task 1 (top), task 2 (middle) and the combined loss (bottom) on the convex toy continual learning problem for different optimization methods.
        {\bfseries (B)}~As in (A), now for the non-convex problem.
        }
    \vspace*{-1em}
\end{figure}

In \CrefNoLink{fig:schematic}A, we consider two regression tasks with losses defined as:
\begin{align}
	\ell_1(\b{\theta}) & = \frac{1}{2} (\b{\theta}-\b{\theta}_1)^T \b{Q}_1 (\b{\theta}-\b{\theta}_1)  \\
	\ell_2(\b{\theta}) & = \frac{1}{2} (\b{\theta}-\b{\theta}_2)^T \b{Q}_2 (\b{\theta}-\b{\theta}_2), 
\end{align}
where $\b{\theta}_1 = (3, -6)^\top$, $\b{\theta}_2 = (3,6)^\top$,
\begin{align}
	\b{Q}_1     & =           
	\b{R}(\phi_1)
	\begin{bmatrix}
	1           & 0           \\
	0           & \zeta       
	\end{bmatrix}
	\b{R}(\phi_1)^T,\\
	\b{Q}_2     & =           
	\b{R}(\phi_2)
	\begin{bmatrix}
	2           & 0           \\
	0           & \zeta       
	\end{bmatrix}
	\b{R}(\phi_2)^T, \\
	\b{R}(\phi) & =           
	\begin{bmatrix}
	\cos(\phi)  & -\sin(\phi) \\
	\sin(\phi)  & \cos(\phi)  
	\end{bmatrix},
\end{align}
and $\zeta = 5.5$.
We `train' on task 1 first by setting $\b{\theta} = \b{\theta}_1$.
We then construct a Laplace approximation to the posterior after learning task 1 to find the posterior precision $\b{Q}_1$ (which is in this case exact since the loss is quadratic in $\b{\theta}$).
Now we proceed to train on task 2 by maximizing the posterior (see ~\CrefNoLink{eq:laplace_update_mu}):
\begin{align}
	\mathcal{L}_2(\b{\theta}) & = \ell_2(\b{\theta}) + \frac{1}{2} (\b{\theta} - \b{\theta}_1)^T \b{Q}_1 (\b{\theta}-\b{\theta}_1) \\
	                     & = \ell_2(\b{\theta}) + \ell_1(\b{\theta})                                           
\end{align}
The gradient of $\mathcal{L}_2(\b{\theta})$ with respect to $\b{\theta}$ is given by:
\begin{equation}
	\nabla_{\b{\theta}} \mathcal{L} = \b{Q}_1 (\b{\theta} - \b{\theta}_1) + \b{Q}_2 (\b{\theta}-\b{\theta}_2).
\end{equation}
We can optimize $\ell(\b{\theta})$ using the following three methods:
\begin{align}
	\text{Laplace:} 
	\quad 
	\Delta \b{\theta}
	  & \propto                          
	\b{Q}_1 (\b{\theta} - \b{\theta}_1) + \b{Q}_2 (\b{\theta} - \b{\theta}_2) \\
	\text{NCL:} 
	\quad 
	\Delta \b{\theta}
	  &                                  
	\propto
	 (\b{\theta} - \b{\theta}_1) 
	+ \b{Q}_1^{-1}\b{Q}_2 (\b{\theta} - \b{\theta}_2) \\
	\text{Projected:} 
	\quad 
	\Delta \b{\theta}
	  &                                  
	\propto
	\b{Q}_1^{-1} \b{Q}_2 (\b{\theta} - \b{\theta}_2),
\end{align}
where $\gamma$ is the learning rate and $\b{Q}_1 + \b{Q}_2$ is the Hessian of $\mathcal{L}(\b{w})$.
Note that in `projected', we optimize on task 2 only rather than on the Laplace posterior.
In \CrefNoLink{fig:schematic}B, we consider a slight modification to $\ell_2$ such that the loss is no longer convex:
\begin{equation}
	\ell_2(\b{w}) = 
	\frac{1}{2} (\b{\theta}-\b{\theta}_2)^T \b{Q}_2 (\b{\theta}-\b{\theta}_2) 
	+ a - 
	a \exp \left ( 
	-\frac{1}{2}
	(\b{\theta}-\b{v})^T \b{Q}_v (\b{\theta}-\b{v})
	\right ),
\end{equation}
where we have added a Gaussian with covariance $\b{Q}_v$ to the second loss.
The NCL preconditioner from task 1 remains unchanged ($\b{Q}_1^{-1}$) since $\ell_1$ is unchanged. 
Denoting $G := a \exp \left ( -\frac{1}{2} (\b{\theta}-\b{v})^T \b{Q}_v (\b{\theta}-\b{v}) \right )$, we thus have the following updates when learning task 2:
\begin{align}
	\text{Laplace:} 
	\quad
	\b{\Delta} \b{\theta}   
	  & \propto 
	\b{Q}_1 (\b{\theta} - \b{\theta}_1) 
	+ \b{Q}_2 (\b{\theta} - \b{\theta}_2) + 
	\b{Q}_v (\b{\theta} - \b{v}) G
	\\ 
	\text{NCL:} 
	\quad 
	\b{\Delta} \b{\theta}
	  & \propto 
	(\b{\theta} - \b{\theta}_1) + 
	\b{Q}_1^{-1} \b{Q}_2 (\b{\theta} - \b{\theta}_2) + 
	\b{Q}_1^{-1} \b{Q}_v (\b{\theta} - \b{v}) G
	\\
	\text{Projected:} 
	\quad 
	\b{\Delta}{\b{\theta}}
	  & \propto 
	  \b{Q}_1^{-1} \b{Q}_2 (\b{\theta} - \b{\theta}_2) + 
	  \b{Q}_1^{-1} \b{Q}_v (\b{\theta} - \b{v}) G.
\end{align}
In this non-convex case, the different methods can converge to different local minima (c.f. \CrefNoLink{fig:schematic}B).

The losses on both tasks as well as the combined loss as a function of optimization step are illustrated in \Cref{fig:schem_losses} for the convex and non-convex settings.

\end{appendices}

\bibliographystyleAPP{apalike}
\bibliographyAPP{appendix/appendix}

\end{document}